\documentclass[a4paper,fleqn]{cas-dc}

\usepackage{amsmath}
\usepackage{caption}
\usepackage{url}
\usepackage{enumitem}
\usepackage{pifont}
\usepackage[T1]{fontenc}        
\usepackage{graphicx}           
\usepackage[left,mathlines]{lineno}   
\usepackage[svgnames]{xcolor}   
\usepackage{colortbl}           
\usepackage{amsmath}            
\usepackage{amssymb}            
\usepackage[numbers, sort&compress]{natbib}
\usepackage{multirow}
\usepackage{booktabs}
\usepackage{hyperref}
\hypersetup{colorlinks=true, linkcolor=red, anchorcolor=blue, citecolor=green, urlcolor=LightSeaGreen}

\usepackage[capitalize]{cleveref}
\crefname{section}{Sec.}{Secs.}
\Crefname{section}{Section}{Sections}
\crefname{table}{Tab.}{Tabs.}
\Crefname{table}{Table}{Tables}
\crefname{figure}{Fig.}{Figs.}
\Crefname{figure}{Figure}{Figures}
\crefname{equation}{Eq.}{Eqs.}
\Crefname{equation}{Equation}{Equations}

\usepackage[numbers]{natbib}

\newcommand{\nohave}{\color{red} \ding{55}}
\newcommand{\have}{\color{green} \ding{51}}
\newcommand{\smallsection}[1]{\noindent \textbf{#1.}}

\definecolor{Gray}{gray}{0.85}
\definecolor{LightBlue}{HTML}{94C3E1}
\definecolor{key_words}{HTML}{ec008d}
\definecolor{annotation}{HTML}{009900}

\usepackage{newfloat}
\usepackage{listings}
\usepackage{xspace}
\makeatletter
\DeclareRobustCommand\onedot{\futurelet\@let@token\@onedot}
\def\@onedot{\ifx\@let@token.\else.\null\fi\xspace}

\def\etal{\emph{et al}\onedot}
\makeatother

\def\tsc#1{\csdef{#1}{\textsc{\lowercase{#1}}\xspace}}
\tsc{WGM}
\tsc{QE}
\begin{document}
\let\WriteBookmarks\relax
\def\floatpagepagefraction{1}
\def\textpagefraction{.001}

\shorttitle{Hyperspectral Image Classification via Efficient Global Spectral Supertoken Clustering}    

\shortauthors{Peifu Liu, et al}  

\title [mode = title]{Hyperspectral Image Classification via Efficient Global Spectral Supertoken Clustering}  



%

\author[1]{Peifu Liu}[orcid=0000-0003-4018-1478]
\ead{3120245389@bit.edu.cn}

\author[1,2,3]{Tingfa Xu}[orcid=0000-0001-5452-2662]
\cormark[1]
\ead{ciom_xtf1@bit.edu.cn}

\author[1]{Jie Wang}[orcid=0000-0002-4847-3697]
\ead{3220245170@bit.edu.cn}

\author[1]{Huan Chen}[orcid=0000-0003-1965-9107]
\ead{3220235096@bit.edu.cn}

\author[1]{Huiyan Bai}[orcid=0009-0000-8037-2338]
\ead{3120255523@bit.edu.cn}

\author[1,3]{Jianan Li}[orcid=0000-0002-6936-9485]
\cormark[1]
\ead{lijianan@bit.edu.cn}

\address[1]{Beijing Institute of Technology, Beijing 100081, China}
\address[2]{Beijing Institute of Technology Chongqing Innovation Center, Chongqing 401135, China}
\address[3]{Key Laboratory of Photoelectronic Imaging Technology and System, Ministry of Education of China, Beijing 100081, China}

\cortext[1]{Corresponding author.}


\begin{abstract}
Hyperspectral image classification demands spatially coherent predictions and precise boundary delineation. Yet prevailing superpixel-based methods face an inherent contradiction: clustering aggregates similar pixels into regions, but the subsequent classifier operates pixel-wise, undermining regional consistency. Consequently, existing approaches do not guarantee region-level, boundary-aligned classification.
To address this limitation, we propose the Dual-stage Spectrum-Constrained Clustering-based Classifier (DSCC), an end-to-end framework that explicitly decouples clustering from classification by first grouping spectral similar and spatially proximate pixels into \textit{spectral supertokens} and then performing token-level prediction.
At its core, DSCC computes an image-level multi-criteria feature distance between pixels and centers, followed by a locality-aware assignment regularization, enabling the generation of boundary-preserving spectral supertokens.
A density-isolation based center selection further yields representative, well-separated centers, reducing redundancy and improving robustness to scale variation.
To accommodate mixed land-cover compositions within each token, we introduce a soft-label scheme that encodes class proportions and improves robustness for mixed-class tokens.
DSCC attains a CF1 of 0.728 at 197.75 FPS on the WHU-OHS dataset, offering a superior accuracy-efficiency trade-off compared with state-of-the-art methods. Extensive experiments further validate the effectiveness and generality of the proposed dual-stage paradigm for hyperspectral image classification.
The source code is available at \url{https://github.com/laprf/DSCC}.
\end{abstract}


\maketitle

\section{Introduction}
Hyperspectral images (HSIs) capture rich spatial and spectral information at high resolution, enabling precise material characterization~\cite{MambaHSI,10677534}. Hyperspectral image classification (HSIC), which assigns each pixel to a predefined category, is fundamental in remote sensing and supports applications such as object detection~\cite{HSOD-BIT-V2}, urban planning~\cite{urban}, agricultural monitoring~\cite{10144690,BHADRA20241}, and mineral exploration~\cite{mining}.

\begin{figure}
    \centering
    \includegraphics[width=\linewidth]{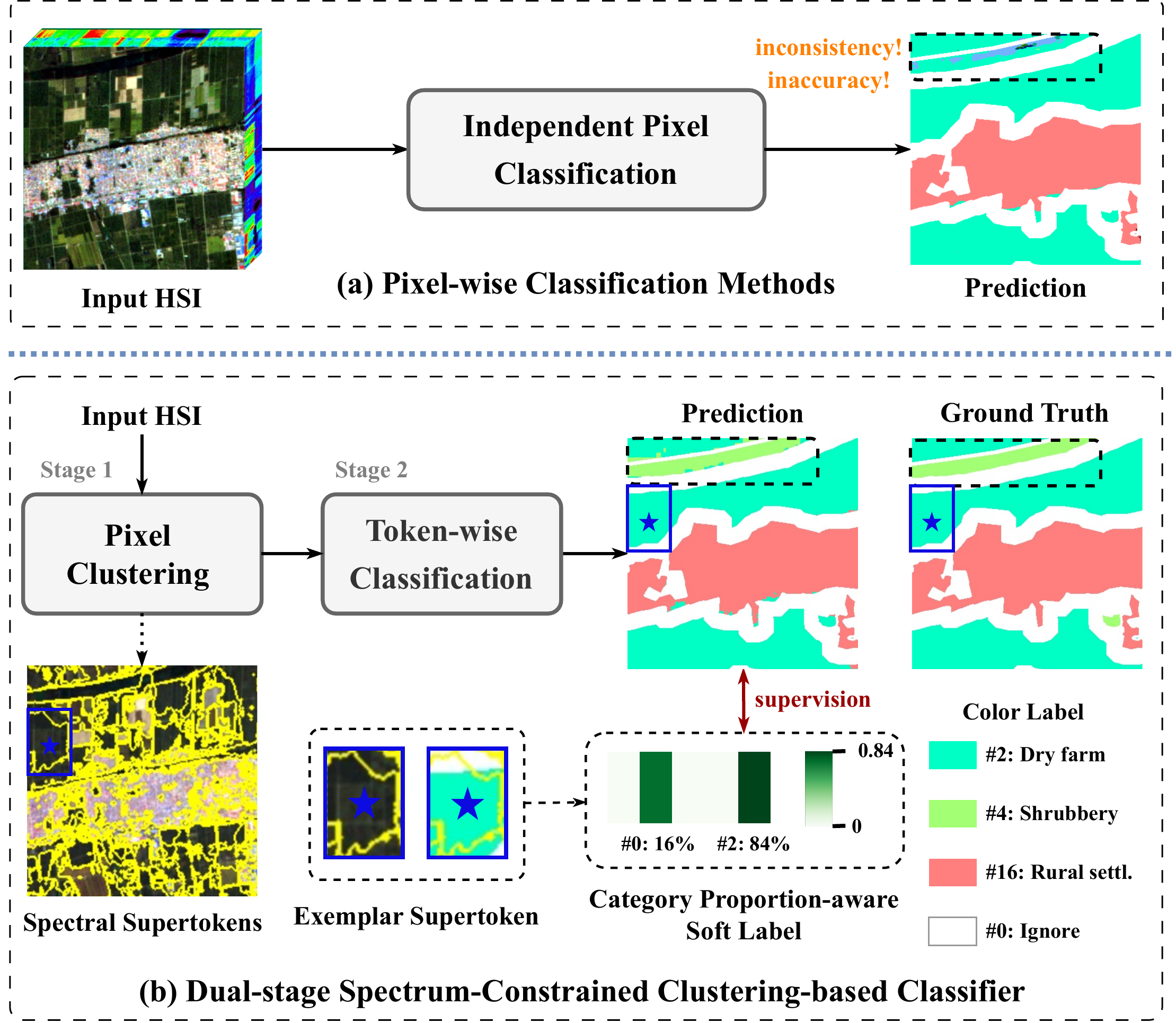}
    \caption{
    \textbf{(a)} Pixel-wise classification models predict each pixel independently and often produce local inconsistencies and inaccurate boundaries (black boxes). Our proposed \textbf{(b)} Dual-stage Spectrum-Constrained Clustering-based Classifier first aggregates spectrally similar and spatially adjacent pixels into \textbf{spectral supertokens} and then classifies them with a category proportion-aware soft label, yielding region-consistent predictions with sharper boundaries.
   }
    \label{fig:motivation}
\end{figure}

Recent deep learning approaches, including convolutional neural networks (CNNs)~\cite{3D-CNN, FreeNet, ReS3Net, CSCN, PANDE2022422}, Transformers~\cite{SpectralFormer, GAHT, YANG2023145}, and Mamba architectures~\cite{MambaHSI, hypermamba}, have significantly advanced classification performance. However, as shown in~\cref{fig:motivation}(a), two limitations persist: \textbf{(i)} pixel-wise classification fails to treat similar pixels as coherent regions, producing local inconsistencies in prediction maps; \textbf{(ii)} fixed-grid mechanisms such as CNN kernels and Transformer patches adapt poorly to complex boundaries, leading to inaccurate edge delineation.

These issues motivate the integration of clustering into hyperspectral image classification. By grouping spectrally similar and spatially adjacent pixels, clustering creates semantically meaningful units that better align with object boundaries. Prior works exploit clustering or superpixels—for instance, to guide deformable convolutions~\cite{Superpixel-Guided} or to construct superpixel-level features~\cite{Superpixel_Contracted}. Yet their final predictions remain pixel-wise. This creates an architectural contradiction: although clustering can capture homogeneous regions during feature extraction, the classifier ultimately treats pixels independently, breaking the spatial coherence induced by clustering. Consequently, local inconsistencies persist, indicating that simply injecting clustering into a pixel-wise pipeline cannot ensure region-consistent classification. True region-wise classification is needed.

Motivated by this, we decouple clustering from classification and propose the \textbf{D}ual-stage \textbf{S}pectrum-Constrained \textbf{C}lustering-based \textbf{C}lassifier (DSCC), an end-to-end dual-stage hyperspectral image classification framework, as shown in~\cref{fig:motivation}(b). In Stage 1, DSCC forms spectral supertokens through global pixel-center association and refines them via center filtering. In Stage 2, these supertokens are classified token-wise and then mapped back to pixels, preserving spatial-spectral consistency throughout the pipeline. This dual-stage design ensures genuinely region-wise rather than pixel-wise prediction.

Implementing this dual-stage paradigm in an end-to-end and efficient manner introduces distinct challenges. Existing clustering mechanisms rely on simple spatial-color metrics such as XYRGB~\cite{Super_sampling_network, Context_cluster, Chen_2024_CVPR}, limiting their ability to exploit spectral information. Our previous work~\cite{DSTC_ECCV} improves spectral modeling but restricts pixel-center similarity computation to \textbf{local} patches. This patch-based design inevitably truncates large, continuous regions at patch boundaries and introduces discontinuities. Moreover, the element-wise affinity computation between centers and neighboring pixels imposes a significant inference bottleneck.

To address these issues, we propose the Spectral-Consistent Pixel Aggregation (SCPA) mechanism. SCPA models \textbf{global} pixel-center similarity using a multi-criteria feature distance that integrates spectral, spatial, and structural cues. Subsequently, pixels that are spectrally similar and spatially adjacent are aggregated to form spectral supertokens. By avoiding fixed patch-wise partition constraints and modeling pixel-center relations in an image-level candidate space, spectral supertokens can better preserve region coherence and alleviate boundary truncation. The parallelized, one-shot computation nature of SCPA also significantly accelerates inference.

Uniformly distributing centers across an image can produce an imbalanced token distribution: large homogeneous regions accumulate excessive centers, generating redundant, overly fragmented tokens, while small or fine-grained structures receive too few, weakening their representation. To mitigate this, we introduce the Density-Isolation Center Filtering (DICF) mechanism, which evaluates centers by their local density and mutual separation. DICF selects representative, well-isolated centers, effectively reduces token redundancy and enhances robustness to scale variation.

A further challenge in dual-stage hyperspectral image classification is supervising region-wise predictions. Even with carefully constructed supertokens, many inevitably cover pixels from multiple classes, especially near boundaries or in mixed land-cover areas. Training with one-hot labels at the token level would introduce severe label noise. To address this, we propose a Category Proportion-aware Soft Label that encodes the class proportions within each supertoken, as illustrated in~\cref{fig:motivation}(b). This supervision aligns with the internal composition of tokens, improves robustness to mixed-class regions, and stabilizes training.

\begin{figure}
    \centering
    \includegraphics[width=\linewidth]{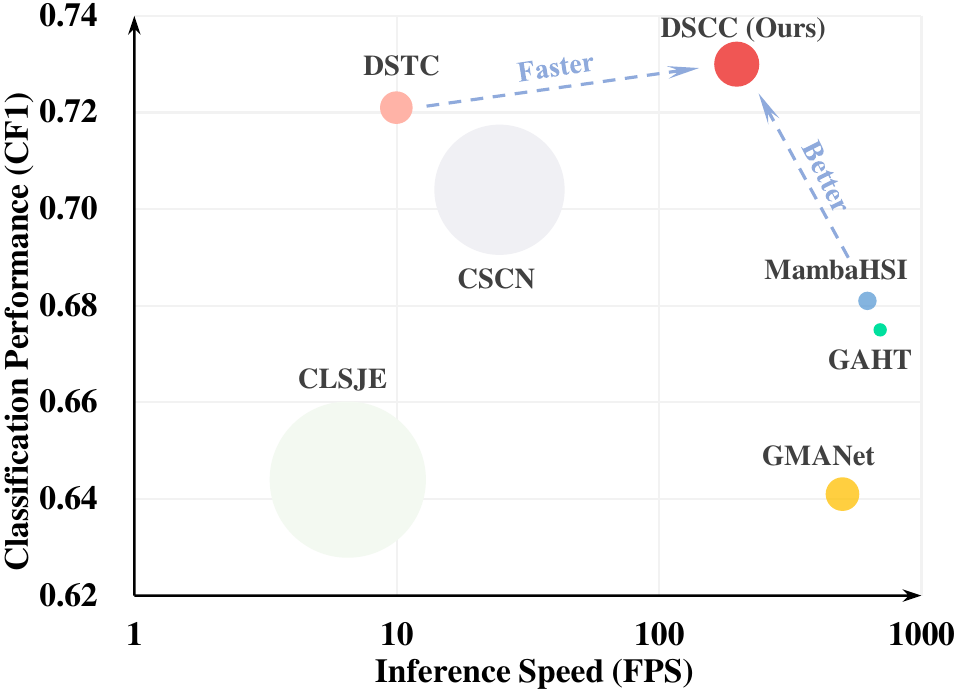}
    \caption{Comparison of classification performance and inference speed. Circle radius represents Floating Point Operations (FLOPs). Our DSCC achieves the best accuracy-efficiency trade-off.}
    \label{fig:cf1_and_fps}
\end{figure}

We evaluate DSCC on the WHU-OHS dataset~\cite{WHU-OHS} and compare it with open-source state-of-the-art methods. As shown in~\cref{fig:cf1_and_fps}, DSCC achieves a CF1 of 0.728 at 197.75 FPS, demonstrating a strong accuracy-efficiency trade-off. Additional experiments on the IP, KSC, WHU-Hi-Hanchuan~\cite{WHU-Hi}, and HyRANK-Dioni datasets further confirm its generality.
This work extends and deepens our previous conference paper~\cite{DSTC_ECCV} by introducing several innovations and addressing core limitations:
\begin{itemize}[leftmargin=*, itemsep=2pt, topsep=0pt, parsep=0pt]
\item We introduce a global multi-criteria feature distance mechanism that replaces rigid patch-based computations with a full-image distance evaluation strategy, maintaining similarity accuracy while boosting inference speed.
\item Building on this distance mechanism, we propose the Spectral-Consistent Pixel Aggregation module, which aggregates spatially related pixels into spectral supertokens. This global search space coupled with dynamic local assignment resolves the boundary truncation inherent in fixed-grid hard aggregation.
\item We design a Density-Isolation Center Filtering mechanism that evaluates both local density and global isolation to filter representative and well-
separated centers, improving robustness to scale variation.
\end{itemize}

\section{Related Work}
\subsection{Superpixel in Hyperspectral Image Classification}
Despite advances enabled by CNNs~\cite{3D-CNN, FreeNet, ReS3Net, CSCN}, Transformers~\cite{SpectralFormer, GAHT}, and Mamba architectures~\cite{MambaHSI, hypermamba}, hyperspectral image classification still suffers from segmentation discontinuities and imprecise edge localization. Superpixel segmentation is a promising way to mitigate these issues. For instance,
Zhao~\etal~\cite{Superpixel-Guided} used superpixels to guide deformable convolutions for feature extraction, while Tu~\etal~\cite{tu_2023_hyperspectral} and Nartey~\etal~\cite{nertey_2023_picovs} integrated superpixels into graph neural networks to enhance boundary modeling. Bai~\etal~\cite{bai_2022_hyperspectral} further combined superpixels with graph-based methods to balance spatial-spectral representation and computational efficiency. However, these approaches still perform pixel-wise classification, leading to local inconsistencies.

Our method addresses this issue by aggregating pixels into spectral supertokens and performing token-wise classification. Pixel-level predictions are obtained by assigning all pixels in a supertoken to its predicted class. This strategy improves edge delineation and regional consistency while remaining efficient and effective for hyperspectral image classification.

\subsection{Transformer/Mamba in Hyperspectral Image Classification} 
Transformers, known for self-attention and strong long-range dependency modeling~\cite{attention}, have been widely adopted in hyperspectral image classification. Hong~\etal~\cite{SpectralFormer} pioneered this line by replacing CNN-based preprocessing with group-wise spectral embeddings processed by a Transformer encoder. Subsequent works refined spectral-spatial modeling: Mei~\etal~\cite{GAHT} proposed a hierarchical Transformer with grouped pixel embeddings; Yu~\etal~\cite{CLSJE} introduced CLSJE, integrating multi-scale features with attention to handle class imbalance; Lu~\etal~\cite{GMANet} designed a band-grouping multi-attention network to extract discriminative spectral-spatial features.
More recently, Mamba~\cite{mamba}, a structured state space model with linear-time complexity and strong long-range modeling capacity, has emerged as an alternative to Transformers. Li~\etal~\cite{MambaHSI} proposed MambaHSI with adaptive spectral-spatial fusion, while Sheng~\etal~\cite{dualmamba} introduced DualMamba, a lightweight parallel CNN-Mamba architecture with cross-attention fusion to capture both global and local cues.

Unlike one-stage Transformer- or Mamba-based approaches, our method adopts a dual-stage design: it first aggregates similar pixels into supertokens and then applies a hybrid Transformer-Mamba architecture to model global relations among these tokens, followed by token-wise classification. This paradigm effectively realizes real region-wise prediction.

\subsection{Dual-stage Methods}
Dual-stage methods have been widely used in object detection~\cite{Faster_RCNN}, where region proposals are first generated and then classified and refined. Recent studies have also shown that dual-stage deep architectures remain effective across diverse vision tasks. For example, stage-wise designs have been adopted for face anti-spoofing~\cite{wang2022disentangled}, where identity-related features are first separated from spoofing cues to improve cross-domain generalization, for human-object interaction detection~\cite{10658050}, where intermediate spatial priors are first constructed and then used for interaction recognition, for motion prediction via proposal generation and subsequent trajectory refinement~\cite{Choi_2023_ICCV}, and for neural surface reconstruction through two-stage representation learning and refinement~\cite{NEURIPS2024_bade1564}.

In hyperspectral image classification, Tu~\etal~\cite{tu2019dual} proposed an early dual-stage framework for generating probability maps. Their method extracted hand-crafted shape attributes to produce an initial classification map and then refined it using a rolling guidance filter. Although effective at preserving boundaries, it suffered from limited accuracy and efficiency due to reliance on manual features. Recent studies have further shown that stage-wise modeling can also be beneficial for hyperspectral classification when subpixel priors are explicitly incorporated. For example, S2VNet~\cite{10856229} first learns subpixel representations through an unmixing-oriented spectral modeling branch and then integrates them with classifier features for final prediction. Although it still produces pixel-wise outputs, this design follows a “subpixel modeling first, semantic decision second” paradigm, suggesting that decoupling intermediate structure-aware representation learning from subsequent classification is beneficial. Motivated by this insight, our method instead formulates a fully region-wise dual-stage framework that explicitly separates supertoken generation and token classification, achieving improved performance and efficiency.

\section{Method}
Let $\boldsymbol{I} \in \mathbb{R}^{\mathrm{H} \times \mathrm{W} \times \mathrm{B}}$ denote the input hyperspectral image, where $\mathrm{H}$ and $\mathrm{W}$ are the spatial dimensions (height and width) and $\mathrm{B}$ is the number of spectral bands. Hyperspectral image classification aims to learn a mapping function $\boldsymbol{f}(\cdot): \mathbb{R}^{\mathrm{H} \times \mathrm{W} \times \mathrm{B}} \rightarrow \mathbb{R}^{\mathrm{H} \times \mathrm{W} \times 1}$ that outputs a pixel-wise classification map:
\begin{equation}
\boldsymbol{Z} = \boldsymbol{f}(\boldsymbol{I}),
\end{equation}
where $\boldsymbol{Z} \in \mathbb{R}^{\mathrm{H} \times \mathrm{W} \times 1}$ is the final classification map.

\begin{figure*}[tp]
    \centering
    \includegraphics[width=\linewidth]{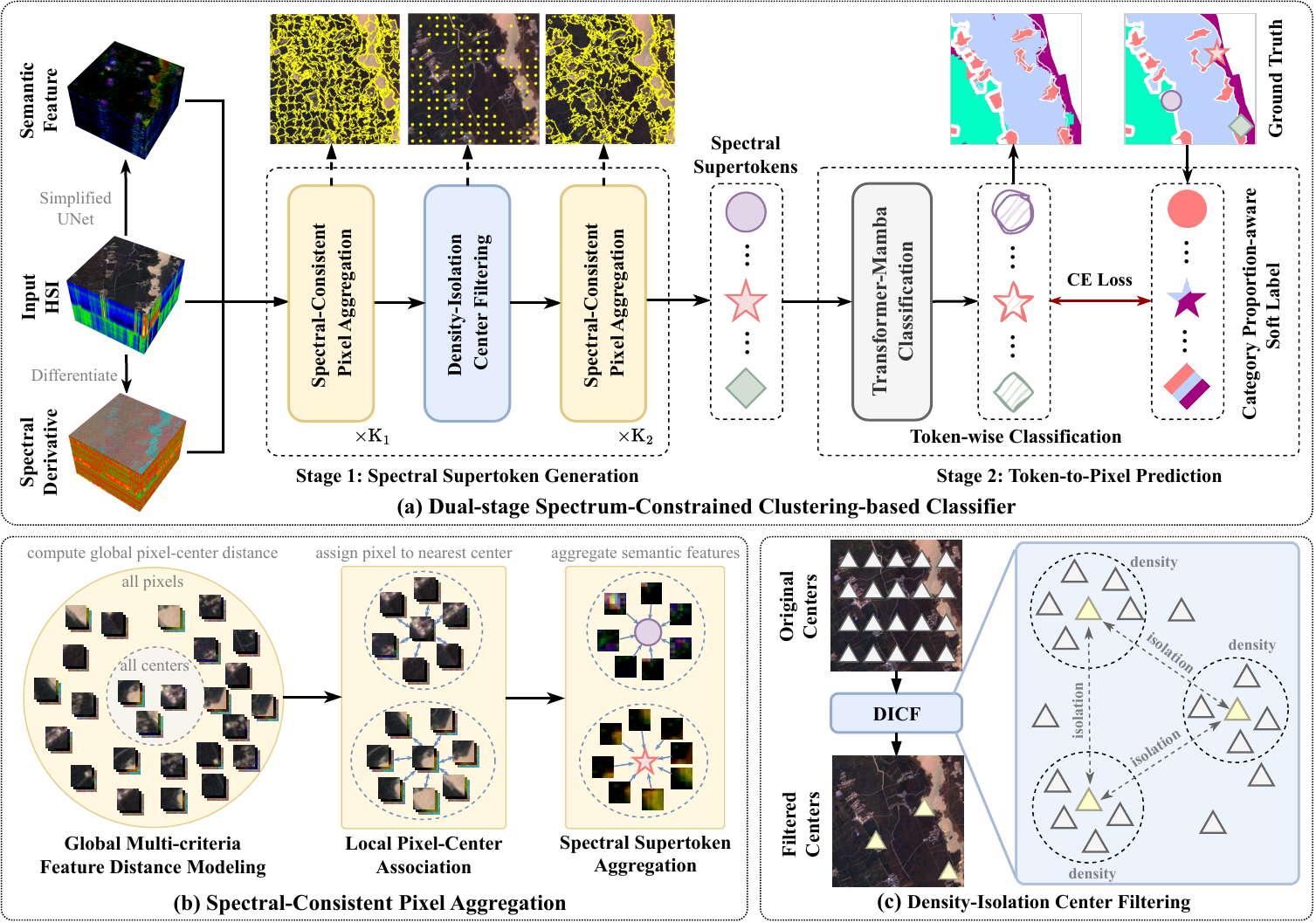}
    \caption{Overall architecture of the proposed \textbf{(a)} Dual-stage Spectrum-Constrained Clustering-based Classifier. In Stage 1, multiple groups of \textbf{(b)} Spectral-consistent Pixel Aggregation is applied to cluster similar pixels into spectral supertokens, followed by \textbf{(c)} Density-Isolation Center Filtering to optimize the distribution of clustering centers. In Stage 2, spectral supertokens are classified using a Transformer-Mamba-based model. Token-wise predictions are projected back to the pixel level to produce the final classification map. During training, each token is supervised by a class proportion-aware soft label, where the color composition indicates the proportion of different land cover types within each token.
    }
    \label{fig:overall}
    \vspace{-4mm}
\end{figure*}

Conventional methods implement $\boldsymbol{f}(\cdot)$ in a strictly pixel-wise manner, leading to local inconsistencies and inaccurate edges. To overcome this, we reformulate hyperspectral image classification as region-wise prediction via a two-stage paradigm: Stage 1 aggregates spectrally similar and spatially proximate pixels into tokens, and Stage 2 performs token-wise classification and maps predictions back to pixels.
To this end, we propose a Dual-stage Spectrum-Constrained Clustering-based Classifier (DSCC), as shown in~\cref{fig:overall}(a).

Specifically, Stage~1 consists of two groups of Spectral-Consistent Pixel Aggregation (SCPA) modules, where each module is repeated $\mathrm{K}_l \, (l = 1,2)$ times. In each group, all pixels are associated with $\mathrm{M}_l  \, (l = 1,2)$ clustering centers in a single global step, producing spectral supertokens. A Density-Isolation Center Filtering (DICF) module between the two SCPA groups selects centers that are both representative and globally isolated. In Stage~2, the refined supertokens are fed into a Transformer-Mamba-based classifier for token-wise classification. Supervision is provided by a Category Proportion-aware Soft Label that encodes the class proportions within each supertoken.

\subsection{Preliminary Spectral-Semantic Analysis}
To support the subsequent modules, we extract both semantic features and spectral derivatives from the hyperspectral image.

\smallsection{Semantic Feature Extraction}
Simply using the original spectra alone may be insufficient for robust clustering, while deep semantic features provide complementary information. In our framework, the semantic feature map \(\boldsymbol{F} \in \mathbb{R}^{\mathrm{H} \times \mathrm{W} \times \mathrm{C}_1}\) is obtained as:
\begin{equation}
    \boldsymbol{F} = \boldsymbol{f}_\text{U}(\boldsymbol{I}),
\end{equation}
where \(\mathrm{C}_1\) denotes the feature dimension, and \(\boldsymbol{f}_\text{U}(\cdot)\) represents the feature extractor. To balance computational efficiency and performance, \(\boldsymbol{f}_\text{U}(\cdot)\) is implemented using a lightweight UNet architecture with two downsampling and two upsampling layers.

\smallsection{Spectral Derivative Computation}
The spectral derivative highlights local spectral structures and suppresses subtle fluctuations~\cite{WAN2021112761}. We incorporate it into our model as an additional cue.
Let $\boldsymbol{I}(i) \in \mathbb{R}^{\mathrm{H} \times \mathrm{W}}$ be the $i$-th spectral band of the input image. The spectral derivative at band $i$, denoted $\boldsymbol{I}^\prime(i) \in \mathbb{R}^{\mathrm{H} \times \mathrm{W}}$, is computed as:
\begin{equation}
    \boldsymbol{I}^\prime(i) = \boldsymbol{I}(i+1) - \boldsymbol{I}(i), \quad i = 1, \dots, \mathrm{B} - 1.
\end{equation}
Applying this computation across all spectral bands yields the spectral derivative \(\boldsymbol{I}^\prime \in \mathbb{R}^{\mathrm{H} \times \mathrm{W} \times (\mathrm{B} - 1)}\).

\subsection{Spectral-Consistent Pixel Aggregation}
Patch-based clustering methods~\cite{DSTC_ECCV, Context_cluster, Chen_2024_CVPR} restrict pixel-center association to local windows. This truncates large homogeneous regions at patch boundaries and causes discontinuities. Iterative, patch-wise updates also introduce computation bottleneck.

To avoid these issues, we adopt a one-shot global association strategy that computes a global multi-criteria feature distance between each pixel and every clustering center over the entire image. As illustrated in~\cref{fig:overall}(b), this process consists of three steps: (i) computing global multi-criteria feature distances between pixels and centers, (ii) enforcing local pixel-center association, and (iii) aggregating spectral supertokens. For clarity, we describe the workflow using the first stage as an example, where $\mathrm{M}_1$ centers are initialized.

\smallsection{Global Multi-criteria Feature Distance Modeling}  
Inspired by Context Cluster~\cite{Context_cluster}, we first place a fixed grid of uniformly distributed cluster centers, which balances accuracy and efficiency. We then compute a distance matrix $\boldsymbol{D} \in \mathbb{R}^{\rm N \times M_1}$ that captures the relationships between pixels and centers using both spatial coordinates and multi-modal features:
\begin{equation}
    \label{eq:distance}
    \boldsymbol{D}(n,m) = \boldsymbol{D}_{\text{spa}}(n,m) + \boldsymbol{D}_{\text{feat}}(n,m), 
\end{equation}
where $\rm N = H \times W$ is the number of pixels and $\rm M_1$ is the number of centers.

The spatial distance $\boldsymbol{D}_{\text{spa}} \in \mathbb{R}^{\rm N \times M_1}$ is computed as:
\begin{equation}
 \label{eq:spatial distance}
 \boldsymbol{D}_{\text{spa}}(n,m) = \frac{\| \boldsymbol{C}_\text{p} (n,:) - \boldsymbol{C}_\text{c} (m,:) \|^2} {\max(\mathrm{H},\mathrm{W})},
\end{equation}
where $\boldsymbol{C}_\text{c} \in \mathbb{R}^{\rm M_1 \times 2}$ and $\boldsymbol{C}_\text{p} \in \mathbb{R}^{\rm N \times 2}$ are the spatial coordinates of the centers and pixels, respectively. Based on these coordinates, the spectral features of the centers $\boldsymbol{I}_\text{c} \in \mathbb{R}^{\rm M_1 \times B}$, their spectral derivatives $\boldsymbol{I}^\prime_\text{c} \in \mathbb{R}^{\rm M_1 \times (B-1)}$, and their semantic features $\boldsymbol{F}_\text{c} \in \mathbb{R}^{\rm M_1 \times C_1}$ are sampled from the original image $\boldsymbol{I}$, the derivatives $\boldsymbol{I}^\prime$, and the semantic feature map $\boldsymbol{F}$, respectively.
The feature distance matrix $\boldsymbol{D}_{\text{feat}} \in \mathbb{R}^{\rm N \times M_1}$ is subsequently defined as:
\begin{equation}
    \label{eq:feature distance}
    \begin{aligned}
 \boldsymbol{D}_{\text{feat}}(n,m) & = \frac{ \|\boldsymbol{I}_\text{r} (n,:) - \boldsymbol{I}_\text{c} (m,:) \|^2}{\sqrt{\mathrm{B}}} \\
  &+ \frac{\| \boldsymbol{I}^\prime_\text{r} (n,:)- \boldsymbol{I}^\prime_\text{c} (m,:)\|^2}{\sqrt{\mathrm{B-1}}}  \\
  &+ \frac{\| \boldsymbol{F}_\text{r} (n,:) - \boldsymbol{F}_\text{c}(m,:) \|^2}{\sqrt{\mathrm{C}}},
  \end{aligned}
\end{equation}
where $\boldsymbol{I}_\text{r} \in \mathbb{R}^{\rm N \times B}$, $\boldsymbol{I}^\prime_\text{r} \in \mathbb{R}^{\rm N \times (B-1)}$, and $\boldsymbol{F}_\text{r} \in \mathbb{R}^{\rm N \times C_1}$ are the reshaped pixel-wise features. To prevent components with larger magnitudes or higher dimensions from dominating the feature distance, explicit normalization is applied to each term in \cref{eq:spatial distance} and \cref{eq:feature distance}. Specifically, we normalize the spatial distance by the image size and scale each feature distance by the square root of its channel dimension before summation. This guarantees that all modalities contribute fairly to the global multi-criteria feature distance.

Compared with patch-based methods, this image-level formulation allows each pixel to evaluate all clustering centers before locality regularization, so spectrally similar and spatially proximate pixels can still be grouped into supertokens even across patch boundaries. It therefore alleviates the fixed patch-wise truncation of candidate centers caused by patch division, while preserving accurate association. Moreover, the parallel single-pass computation significantly improves efficiency.

\smallsection{Local Pixel-Center Association}
We derive the association matrix $\boldsymbol{A} \in \mathbb{R}^{\mathrm{N} \times \mathrm{M}_1}$, which encodes the relationship between each pixel and all centers, by applying an exponential kernel to the distance matrix:
\begin{equation}
 \boldsymbol{A} = \exp(-\boldsymbol{D}).
\end{equation}
Directly using $\boldsymbol{A}$ may produce spatially scattered assignments, since spectrally similar pixels can be far apart and form fragmented, non-contiguous supertokens. To ensure spatial coherence and prevent fragmented supertokens, we further regularize the image-level association matrix with a spatial top-k mask, which preserves only the nearest candidate centers for each pixel rather than imposing fixed patch-wise isolation.

\smallsection{Spectral Supertoken Aggregation}
We then aggregate the semantic features of all pixels associated with each center, weighted by their similarities.
Let $\boldsymbol{p} \in \mathbb{R}^{\mathrm{C}_1}$ be a center with $\mathrm{N}_\text{p}$ associated feature points, and let $\boldsymbol{a} \in \mathbb{R}^{\mathrm{N}_\text{p}}$ be the corresponding association weights. The aggregated feature $\boldsymbol{s} \in \mathbb{R}^{\mathrm{C}_1}$ is computed as:
\begin{equation}
 \boldsymbol{s} = \frac{\boldsymbol{p} + \sum_{i=1}^{\mathrm{N}_\text{p}} \boldsymbol{a}_i \boldsymbol{f}_{i}}{1 + \sum_{i=1}^{\mathrm{N}_\text{p}} \boldsymbol{a}_i},
\end{equation}
where $\boldsymbol{f}_{i}$ is the $i$-th deep semantic feature. Applying this aggregation to all centers yields the final set of spectral supertokens $\boldsymbol{S}_1 \in \mathbb{R}^{\mathrm{M}_1 \times \mathrm{C}_1}$. These updated center representations are then fed into the Density-Isolation Center Filtering module for center evaluation and selection.

\subsection{Density-Isolation Center Filtering}
Uniformly distributed centers oversample large homogeneous regions and undersample small structures. To reduce token redundancy and improve the adaptability to objects of different scales, we introduce a Density-Isolation Center Filtering (DICF) module, as shown in~\cref{fig:overall}(c), inspired by recent token fusion approaches~\cite{CTM_PAMI}. By jointly assessing local density and global isolation, DICF selects centers that are both representative within their local neighborhoods and well separated in the global feature space.

Unlike conventional methods that compute pairwise distances and merge similar tokens into prototypes, our method differs in two key aspects. First, we use the global multi-criteria feature distance to capture richer relationships among centers in the feature space. Second, we retain the selected centers in their original form instead of merging them, preserving their distinct characteristics and avoiding information loss.

\smallsection{Local Center Compactness}
After spectral supertoken aggregation, each center is represented by an updated feature vector that summarizes the information of its associated pixels. Based on these updated center representations and their corresponding spatial coordinates, we compute the center-to-center distance matrix $\boldsymbol{D}_\text{c} \in \mathbb{R}^{\mathrm{M}_1 \times \mathrm{M}_1}$. Similar to \cref{eq:distance}, this distance still consists of a spatial distance term and a multi-feature distance term. The local density $\rho(j)$ of the $j$-th center is defined based on its $\mathrm{K}$-nearest neighbors:
\begin{equation}
    \rho(j) = \exp\left(-\frac{1}{\mathrm{K}} \sum_{k=1}^\mathrm{K} d^2(j, k)\right) \in(0,1], \quad j \in [1, \mathrm{M}_1],
\end{equation}
where $d(j, k)$ is the distance from the $j$-th center to its $k$-th nearest neighbor, derived from $\boldsymbol{D}_\text{c}$. The local density $\rho(j)$ reflects how compact the neighborhood around the $j$-th center is.

\smallsection{Global Center Separation}
To measure how isolated each center is from others, we define a binary indicator matrix $\boldsymbol{O} \in \mathbb{R}^{\mathrm{M}_1 \times \mathrm{M}_1}$ as:
\begin{equation}
 \boldsymbol{O}(j, k) =
 \begin{cases}
 1, & \text{if } \rho(k) > \rho(j), \\
 0, & \text{otherwise}.
 \end{cases}
\end{equation}
Here, $j$ and $k$ index different centers. For each center, the isolation score $\eta(j)$ is defined as the minimum distance to any other center with higher local density. For the center with the highest density, $\eta(j)$ is set to the maximum distance to all other centers. Formally,
\begin{equation} \small
 \eta(j) = \min_{k} \{ \boldsymbol{O}(j, k) \odot \boldsymbol{D}_\text{c}(j, k) + (1 - \boldsymbol{O}(j, k)) \odot \boldsymbol{D}_{\max} \} \in (0, \boldsymbol{D}_\text{max} ],
\end{equation}
where $\boldsymbol{D}_{\max}$ is the largest value in $\boldsymbol{D}_\text{c}$, and $\odot$ denotes element-wise multiplication. The isolation score quantifies how far the $j$-th center is from denser regions in the feature space.

\smallsection{Joint Score for Center Filtering}
High density ensures that a center is representative of a dense local region, while high isolation ensures that it is distinguishable from centers in other regions. Centers with both high density and high isolation are thus the most representative.

Accordingly, we define a joint score for the $j$-th center as the product of its density and isolation scores:
\begin{equation}
 \text{Score}(j) = \rho(j) \times \eta(j)  \in (0, \boldsymbol{D}_\text{max} ].
\end{equation}
Since $d(j,k)$ is derived from the distance matrix $\boldsymbol{D}_\text{c}$, we have $d(j,k)\ge 0$. Therefore, $\rho(j)\in(0,1]$, $\eta(j)\in(0,\mathbf D_{\max}]$, and $\text{Score}(j)\in(0,\mathbf D_{\max}]$. The final set of centers is obtained by selecting those with the highest scores. This retains only the most distinctive and well-separated centers.

\smallsection{Center Separation Loss}
To further enhance the discriminability of the centers, we introduce a center separation loss. Let $\boldsymbol{C}_\text{e} \in \mathbb{R}^{\mathrm{M_2 \times \mathrm{C}_1}}$ denote the selected centers, where $\mathrm{M}_2$ is the number of selected centers. The average pairwise Euclidean distance between distinct centers is defined as:

\begin{equation}
    \mathrm{d}_\text{e} = \frac{\sum_{i=1}^{\mathrm{M}_2} \sum_{j=1, j \neq i}^{\mathrm{M}_2} \| \boldsymbol{C}_\text{e}(i) - \boldsymbol{C}_\text{e}(j) \|_2}{\mathrm{M}_2(\mathrm{M}_2-1)},
\end{equation}
and the separation loss is defined as $\mathcal{L}_\text{sst} = 1/\mathrm{d}_\text{e}$. This encourages the filtered centers to be well separated in the embedding space, leading to a clearer clustering structure.

\subsection{Token-to-Pixel Prediction}
After filtering, the second group of SCPA modules updates the supertokens, yielding $\boldsymbol{S}_2 \in \mathbb{R}^{\mathrm{M}_2 \times \mathrm{C}_1}$. We subsequently leverage the Transformer’s strength in global feature modeling and Mamba’s efficiency in sequential modeling to perform class prediction for each spectral supertoken:
\begin{equation}
    \hat{\boldsymbol{S}} = \boldsymbol{f}_\text{tm}(\boldsymbol{S}_2),
\end{equation}
where $\boldsymbol{f}_\text{tm}(\cdot)$ denotes a stack of alternating Transformer and Mamba blocks. The output $\hat{\boldsymbol{S}} \in \mathbb{R}^{\rm M_2 \times C}$, where $\mathrm{C}$ is the number of predefined classes.

In our framework, spectral supertokens are fed directly into the model, avoiding the patch partitioning step in Vision Transformers (ViT)~\cite{ViT}. Each Transformer block applies standard self-attention to model token interdependencies. For example, in the first attention block, the self-attention operation is computed as:
\begin{equation}
    \boldsymbol{S}^\prime = \boldsymbol{\sigma}(\frac{\boldsymbol{Q}_\text{S} \boldsymbol{K}^\top_\text{S}}{\sqrt{\mathrm{C}_1}}) \boldsymbol{V}_\text{S},
\end{equation}
where $\boldsymbol{\sigma}(\cdot)$ is the Softmax function, and the \textit{query} $\boldsymbol{Q}_\text{S}$, \textit{key} $\boldsymbol{K}_\text{S}$, and \textit{value} $\boldsymbol{V}_\text{S}$ are obtained via linear projections:
\begin{equation}
    \boldsymbol{Q}_\text{S} = \boldsymbol{S} \boldsymbol{E}_\text{Q}, 
    \boldsymbol{K}_\text{S} = \boldsymbol{S} \boldsymbol{E}_\text{K}, 
    \boldsymbol{V}_\text{S} = \boldsymbol{S} \boldsymbol{E}_\text{V},
\end{equation}
with $\boldsymbol{E}_\text{Q}$, $\boldsymbol{E}_\text{K}$, and $\boldsymbol{E}_\text{V}$ being learnable projection matrices. For brevity, we omit multi-head attention.

Following the Transformer block, the Mamba block applies dynamic state-space modeling to the token sequence. At each time step $t$, given input token $\boldsymbol{S}_t^\prime$, the hidden state $\boldsymbol{h}_t$ is updated as:
\begin{equation}
    \boldsymbol{h}_t = \boldsymbol{f}_\text{A}(\boldsymbol{S}_t^\prime; \theta_\text{A}) \boldsymbol{h}_{t-1} + \boldsymbol{f}_\text{B}(\boldsymbol{S}_t^\prime; \theta_\text{B}) \boldsymbol{S}_t^\prime,
\end{equation}
where $\boldsymbol{f}(\cdot, \theta)$ are learnable functions parameterized by $\theta$, implemented with MLPs to generate state transition and input mapping matrices. The output is then computed as:
\begin{equation}
    \boldsymbol{S}^{\prime\prime}_t = \boldsymbol{C} \boldsymbol{h}_t + \boldsymbol{D} \boldsymbol{S}_t^\prime,
\end{equation}
with $\boldsymbol{C}$ and $\boldsymbol{D}$ are learnable projection matrices.

By alternately stacking Transformer and Mamba blocks, the model captures long-range dependencies and efficiently models sequential behavior. After passing through the stacked blocks, token-wise predictions $\hat{\boldsymbol{S}}$ are obtained via a final linear projection and then mapped back to the image domain to generate the pixel-level classification map $\boldsymbol{Z}$.

\subsection{Category Proportion-aware Soft Label}
In hyperspectral imaging, the ubiquitous mixed-pixel phenomenon gives rise to complex spectral variability, which is often characterized using spatial-spectral or low-rank priors~\cite{9913829}. Analogously, a generated supertoken can be viewed as a macroscopic mixed unit that may encompass multiple classes. To better reflect this compositional property and avoid the label noise introduced by one-hot assignment, we design a Category Proportion-aware Soft Label (Fig. 3). Specifically, it encodes the class proportions within each supertoken, thereby providing supervision that is better aligned with its internal semantic composition.

To handle the multi-class nature of each supertoken, we design a category proportion-aware soft label, as shown in~\cref{fig:overall}. Each pixel is first assigned to its most similar supertoken, and then class proportions are computed within each supertoken’s assigned pixel set.

Specifically, we first determine a hard association matrix $\boldsymbol{A}^\prime \in \{0, 1\}^{\rm N \times M_2}$, where each pixel $n$ is exclusively assigned to the center $m$ with the highest similarity:
\begin{equation}
    \boldsymbol{A}^\prime_{n,m} = 
        \begin{cases}
            1, & \text{if } m = \arg\max_{m'} \boldsymbol{A}_{n,m'} \\
            0, & \text{otherwise}
        \end{cases}
\end{equation}

Given the pixel-level ground truth labels $\boldsymbol{G} \in \{1, ..., \mathrm{C}\}^\mathrm{N}$, we compute the count $\boldsymbol{C}_\text{c}(m,c)$ of class $c$ within supertoken $m$ as:
\begin{equation}
    \boldsymbol{C}_\text{c}(m,c) = \sum_{n=1}^{N} \boldsymbol A^\prime_{n,m} \cdot \delta(\boldsymbol G_n, c),
\end{equation}
where $\delta(x, c)$ is the Kronecker delta function (1 if $x = c$, 0 otherwise).

We then normalize the counts for each supertoken to obtain the soft label matrix $\boldsymbol L \in \mathbb{R}^{\rm M_2 \times C}$:
\begin{equation}
    \boldsymbol L(m,c) = \frac{\boldsymbol C_\text{c}(m,c)}{\sum_{c'=1}^{C} \boldsymbol C_\text{c}(m,c')}.
\end{equation}
The cross-entropy loss is then computed as:
\begin{equation}
    \mathcal{L}_\text{CE}(\hat{\boldsymbol{S}}, \boldsymbol{L}) = -\frac{1}{\mathrm{M}_2} \sum_{m=1}^{\mathrm{M}_2} \sum_{c=1}^{\mathrm{C}} \boldsymbol{L}(m,c) \log \hat{\boldsymbol{S}}(m,c),
\end{equation}
where $\hat{\boldsymbol{S}}(m,c)$ is the predicted probability that supertoken $m$ belongs to class $c$. This soft-label strategy improves robustness to mixed-class tokens and stabilizes training.

\subsection{Learning Objective}
The overall training objective is defined as:
\begin{equation}
\mathcal{L} = \mathcal{L}_\text{CE} + \mathcal{L}_\text{sst},
\end{equation}
where $\mathcal{L}_\text{CE}$ is the cross-entropy loss for classification, and $\mathcal{L}_\text{sst}(l)$ is the supertoken separation loss for the $l$-th module.

\section{Experiments}
We conduct comprehensive comparisons and ablation studies on multiple hyperspectral image classification datasets to evaluate the effectiveness of DSCC and its components.

\subsection{Implementation Details}
We optimize network parameters using the AdamW optimizer with a cosine-annealing learning rate schedule. All experiments are implemented in PyTorch and run on an NVIDIA RTX 3090 GPU with an Intel XEON Gold 5218R CPU. Mixed-precision training is adopted to reduce memory usage and accelerate computation.
The two hierarchy levels contain $\mathrm{K}_1 = 3$ and $\mathrm{K}_2 = 4$ Spectral-Consistent Pixel Aggregation blocks, respectively. In the Local Pixel-Center Association stage, the mask size is set to 9, meaning that each pixel can associate only with its 9 nearest clustering centers. The value of $\mathrm{K}$ in the Density-Isolation Center Filtering module is also set to 9.

\subsection{Results on WHU-OHS Dataset}
\smallsection{Dataset and Experimental Setups}
The WHU-OHS dataset~\cite{WHU-OHS} contains 7795 hyperspectral images acquired by the Orbita Hyperspectral Satellite and covers 24 land cover categories. Each image has a spatial size of 512$\times$512 pixels, a spatial resolution of 10 m, and 32 spectral bands spanning 466-940 nm. As a large-scale hyperspectral benchmark with relatively high spatial detail and complex land-cover boundaries, it is particularly suitable for evaluating region-consistent and boundary-preserving classification methods. During training, the initial learning rate is set to $1 \times 10^{-4}$, with 150 epochs and a batch size of 12. To reduce memory consumption, input images are resized to 256$\times$256 pixels. In the two-level Spectral-Consistent Pixel Aggregation modules, pixels are grouped into $\mathrm{M}_1 = 256$ and $\mathrm{M}_2 = 128$ spectral supertokens at the first and second levels, respectively.

\begin{table*}[ht] 
    \caption{Comparison of DSCC's Performance and Efficiency with Other Methods on WHU-OHS Dataset. CF1: Class Average F1, OA: Overall Average, $\kappa$: Kappa, mIoU: mean Intersection over Union.}
    \label{tab:compare and efficiency}
    \centering
    \setlength{\tabcolsep}{11pt}
    \begin{tabular}{l | c c c c | c c c}
        \toprule[1.2pt]
        Methods & CF1 $\uparrow$ & OA $\uparrow$ & $\kappa$ $\uparrow$ & mIoU $\uparrow$ & FLOPs (G) & \#Params (M) & Speed (FPS) \\
        \midrule
        3D-CNN~\cite{3D-CNN} & - & 0.652 & 0.605 & 0.405 & 1814.51 & 0.346 & 1122.88 \\
        FreeNet~\cite{FreeNet} & - & 0.778 & 0.749 & 0.525 & 116.20 & 2.508 & 231.17 \\
        ViT~\cite{ViT} & 0.673 & 0.774 & 0.745 & 0.542 & 142.08 & 10.049 & 604.15 \\
        CLSJE~\cite{CLSJE} & 0.644 & 0.781 & 0.753 & 0.520 & 412.20 & 5.352 & 6.51 \\
        GAHT~\cite{GAHT} & 0.675 & 0.776 & 0.748 & 0.542 & 3.01 & 0.734 & 696.35 \\
        GMANet~\cite{GMANet} & 0.641 & 0.757 & 0.725 & 0.512 & 19.14 & 1.17 & 550.24 \\
        MambaHSI~\cite{MambaHSI} & 0.681 & 0.776 & 0.748 & 0.546 & 5.746 & 0.170 & 622.66 \\
        CSCN~\cite{CSCN} & 0.704 & 0.788 & 0.761 & 0.574 & 286.50 & 7.38 & 24.65 \\
        3D-CNN+ReS3Net~\cite{ReS3Net} & - & 0.669 & 0.623 & 0.421 & - & - & - \\
        FreeNet+ReS3Net~\cite{ReS3Net} & - & 0.794 & 0.768 & 0.545 & - & - & - \\
        SiT~\cite{SiT} & 0.690 & 0.781 & 0.752 & 0.556 & 45.67 & 44.22 & 333.24 \\
        HDHN~\cite{HDHN} & 0.710 & 0.794 & 0.775 & 0.581 & 50.58 & 8.64 & 92.00 \\
        DSTC~\cite{DSTC_ECCV} & 0.718 & 0.799 & 0.774 & 0.588 & 19.78 & 11.589 & 9.61 \\
        S2Mamba~\cite{S2Mamba} & 0.723 & 0.800 & 0.774 & 0.592 & 139.05 & 7.49 & 501.11 \\
        \midrule
        \rowcolor{Gray}
        DSCC (Ours) & \textbf{0.728} & \textbf{0.802} & \textbf{0.776} & \textbf{0.602} & 34.33 & 8.57 & 197.75 \\
        \bottomrule[1.2pt]
    \end{tabular}
\end{table*}

\begin{figure*}[tp]
    \centering
    \includegraphics[width=\linewidth]{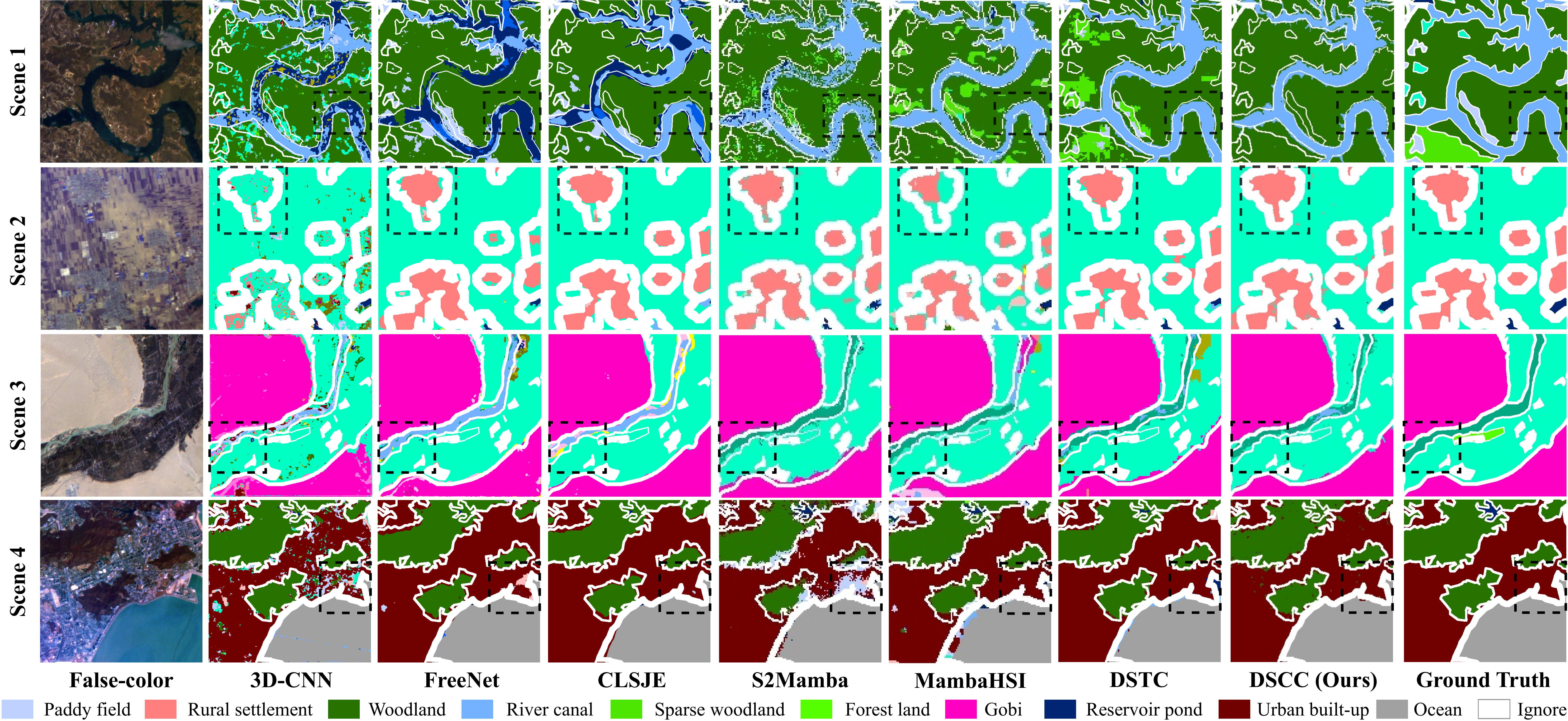}
    \caption{Qualitative results on WHU-OHS. DSCC achieves the best classification accuracy with regional consistency (black-box regions) and precise boundaries.}
    \label{fig:result-vis}
\end{figure*}

\smallsection{Quantitative Results}
\cref{tab:compare and efficiency} reports quantitative comparisons between DSCC and other open-source state-of-the-art methods for hyperspectral image classification. Our previous work DSTC achieves strong performance by aggregating spectrally similar pixels, enabling accurate boundary delineation and region-consistent classification. However, its patch-based clustering strategy introduces boundary truncation artifacts.

To overcome this issue, we enhance the clustering process with a global pixel-center similarity computation that captures long-range dependencies across the entire image, alleviating discontinuities caused by local partitioning. We further introduce Density-Isolation Center Filtering to refine clustering centers, mitigating the uneven supertoken distribution caused by uniform initialization. DSCC achieves consistent gains across multiple metrics. In particular, it attains a CF1 score of 0.728, outperforming DSTC, which achieves 0.718. The mIoU of DSCC reaches 0.602, surpassing DSTC’s 0.588.

\begin{figure}
    \centering
    \includegraphics[width=\linewidth]{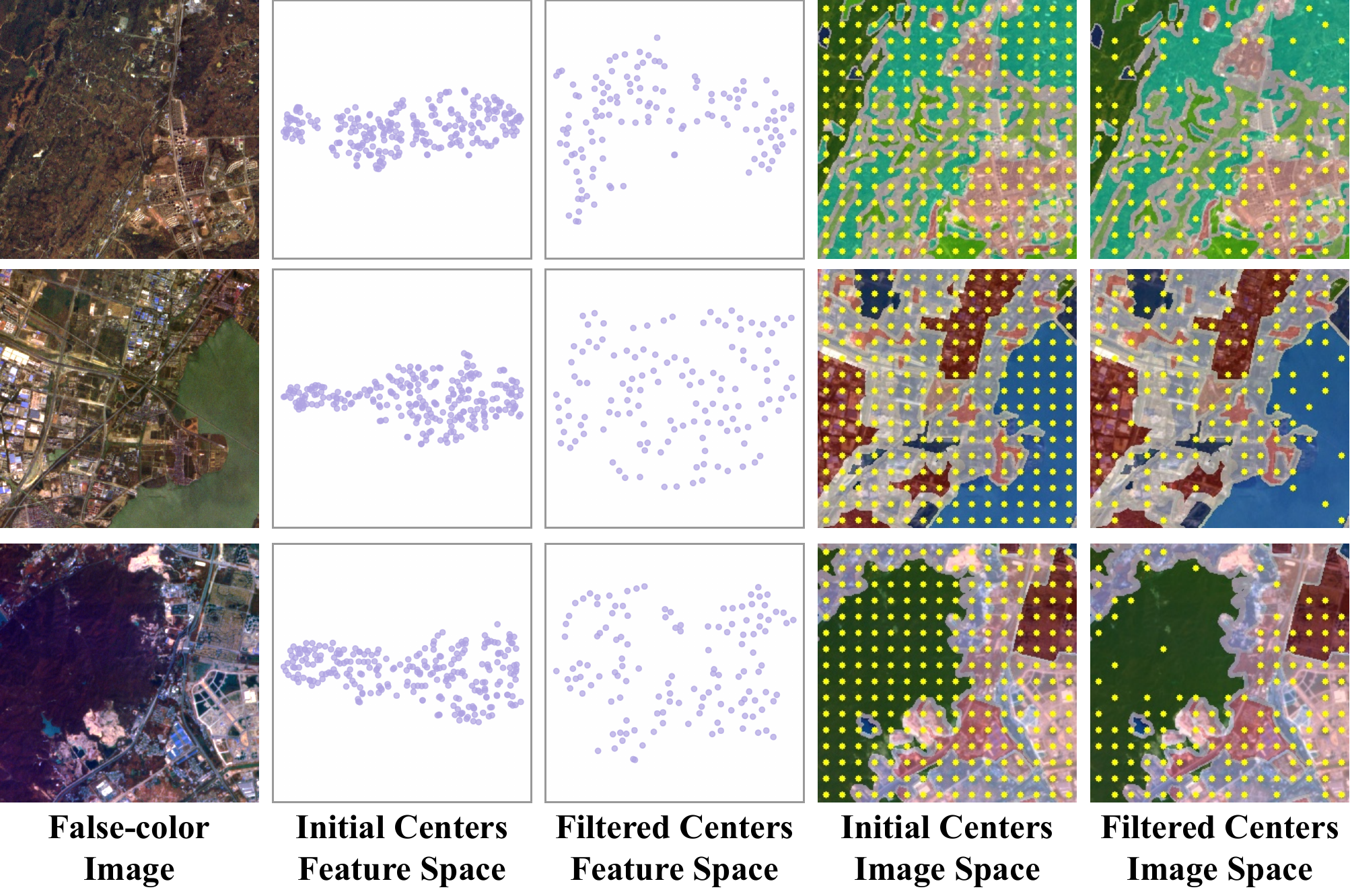}
    \caption{Distribution of clustering centers in feature and image space before and after filtering. Density-Isolation Center Filtering removes less representative centers, resulting in a more uniform and dispersed distribution in the feature space. In image space, centers are reduced in large-scale regions.}
    \label{fig:center-vis}
\end{figure}

Some methods, such as GMANet, MambaHSI, and ReS3Net variants, are specifically tailored for small-scale datasets with limited training samples. On a large-scale dataset like WHU-OHS, where both data volume and image size are large, their performance lags behind CSCN, DSTC, and DSCC. Overall, these results confirm the effectiveness of DSCC in addressing key challenges in hyperspectral image classification.

\smallsection{Efficiency Analysis}
We compare the computational efficiency of DSCC with existing methods in terms of Floating Point Operations (FLOPs), number of parameters (\#Params), and inference speed (FPS). The results are summarized in \cref{tab:compare and efficiency}.
Although DSTC delivers strong classification accuracy, its inference speed is relatively low at about 10 FPS, limiting its suitability for real-time applications. In contrast, DSCC abandons DSTC’s fixed-grid, local aggregation strategy and adopts a one-shot, global paradigm for modeling pixel-center associations, increasing the inference speed to 197.75 FPS.

Despite a moderate increase in model complexity (8.57 M parameters, 34.33 G FLOPs), DSCC achieves the best classification performance among all compared methods on WHU-OHS. While it is not as lightweight as GAHT or MambaHSI, the performance-efficiency trade-off is favorable and justifies the additional computational cost.

\begin{figure}
    \centering
    \includegraphics[width=\linewidth]{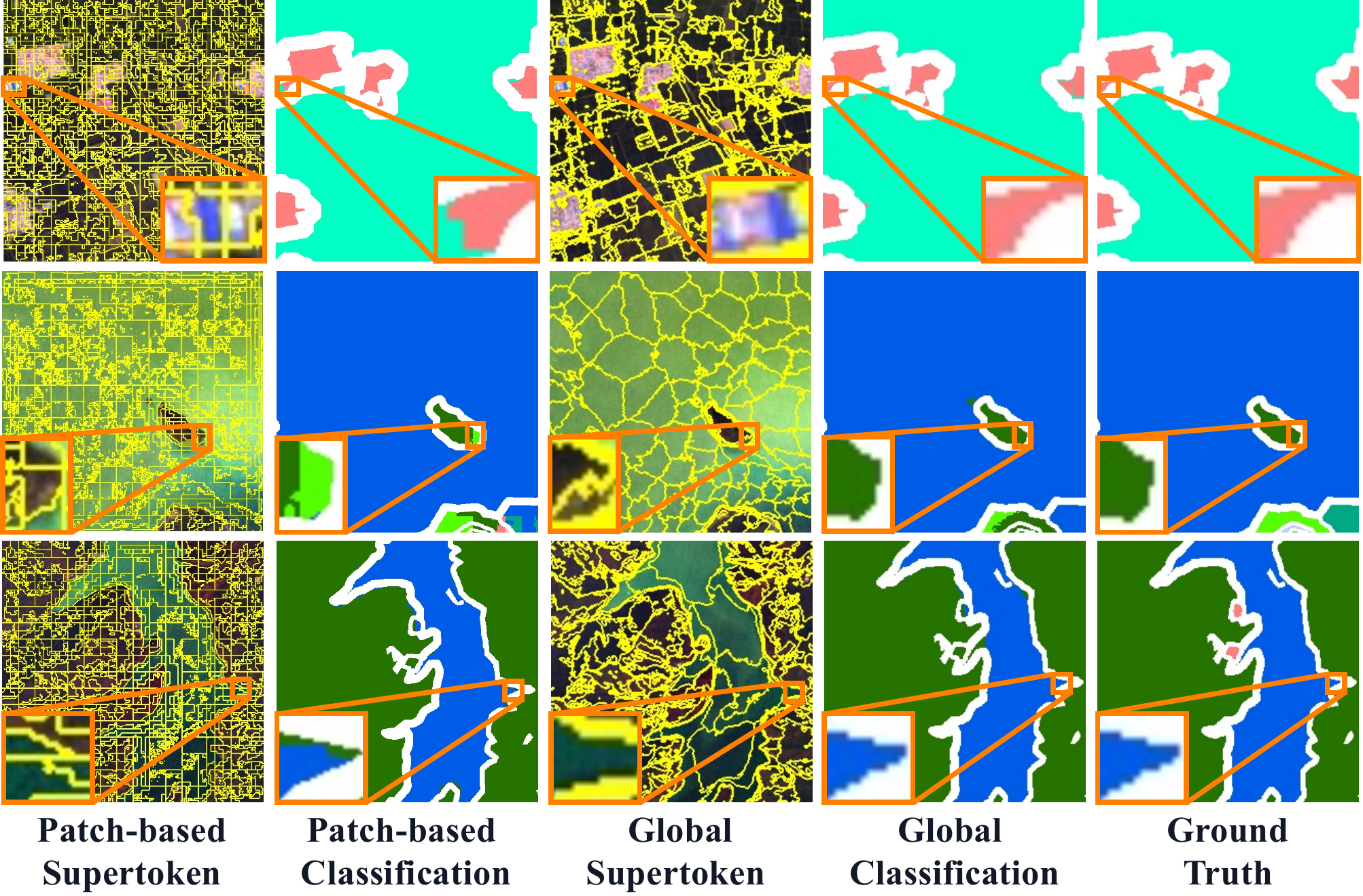}
    \caption{Compared with the patch-based supertoken aggregation method. Our global supertoken aggregation method avoids the truncation caused by patch division at boundaries, thereby preserving structure integrity and delineates accurate boundaries. }
    \label{fig:patch-global-compare}
\end{figure}

\smallsection{Qualitative Results}
As shown in \cref{fig:result-vis}, qualitative results show that DSCC produces more accurate boundaries and better spatial consistency over large regions. The highlighted areas indicate that DSCC yields cleaner and more continuous predictions, such as river bends in Scene~1 and rural settlements in Scene~2. These improvements arise from the use of spectral supertokens, which aggregate spectrally similar and spatially proximate pixels and enable token-wise classification that enhances spatial coherence.

Compared with DSTC, which aggregates pixels within fixed patches, DSCC computes correlations between all pixels and clustering centers globally, and then applies a spatial mask to assign each pixel to the most appropriate cluster. This design helps DSCC better preserve fine-grained boundaries while maintaining the integrity of large regions.

\smallsection{Visualization of Filtered Centers}
\cref{fig:center-vis} visualizes the distribution of clustering centers in both feature and image space before and after applying Density-Isolation Center Filtering. Initially, centers are densely packed in feature space and uniformly distributed in image space. After filtering, only centers that are local density maxima and well separated from other high-density clusters are retained. The remaining centers become more evenly and sparsely distributed in feature space, indicating reduced redundancy. In image space, fewer centers appear in large uniform areas (\textit{e.g.}, woodland), which suppresses redundancy and improves robustness to scale variations.

\smallsection{Comparison with Patch-based Supertoken} 
\cref{fig:patch-global-compare} compares the supertoken segmentation and classification results of the patch-based aggregation in DSTC and the global aggregation in DSCC. Although DSTC can group pixels into supertokens within local patches, it struggles with large or continuous regions, often splitting coherent areas into multiple supertokens. This introduces boundary truncation and imprecise edges, leading to classification errors.

In contrast, our DSCC computes spectral similarity between pixels and clustering centers across the entire image, while a spatial mask restricts pixel assignment to nearby centers. This global strategy better preserves spatial structures and edge continuity, yielding more accurate and consistent classification maps.

\begin{figure}
    \centering
    \includegraphics[width=1\linewidth]{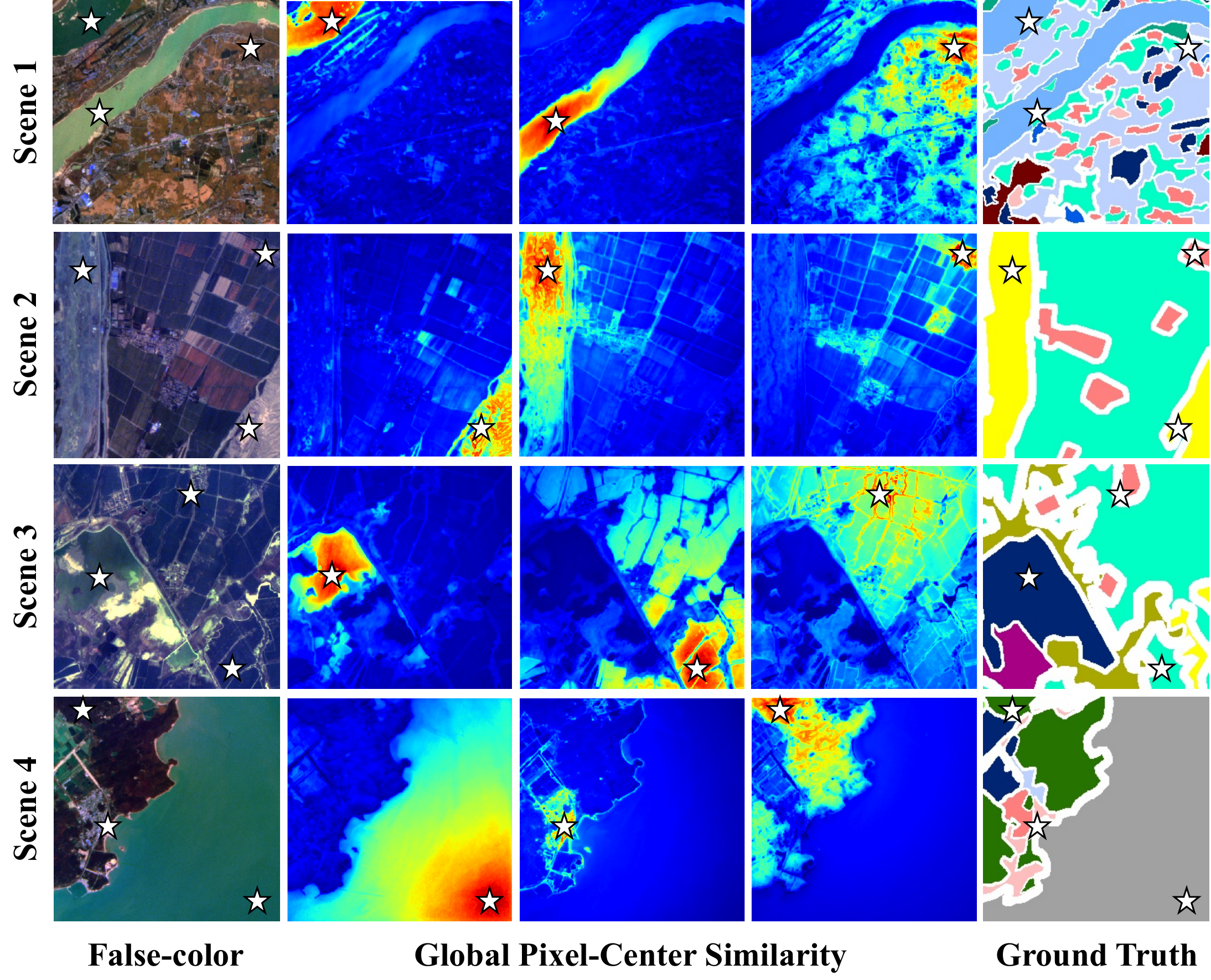}
    \caption{Heatmap of pixel-center similarity. The white stars indicate selected centers. Strong correlations (red) are established with neighboring pixels having similar spectral features, while distant or spectrally distinct pixels show weak correlations (blue).}
    \label{fig:sim mat vis}
\end{figure}

\smallsection{Heatmap of Pixel-Center Similarity}
\cref{fig:sim mat vis} shows the pixel-center similarity in the final Spectral-Consistent Pixel Aggregation block. White stars mark the selected clustering centers, around which strong correlations emerge with neighboring pixels that share similar spectral characteristics. Pixels farther away or belonging to different land cover classes exhibit weaker correlations. This behavior is driven by our Global Multi-criteria Feature Distance, which jointly models spatial and spectral relationships between pixels and centers. Accurate modeling of these correlations enables precise pixel aggregation, thereby improving classification consistency and boundary accuracy.

\subsection{Experiment on Four Benchmark Datasets}

\begin{table*}[tp]
\centering \footnotesize
\setlength{\tabcolsep}{6.4pt}
\caption{Quantitative results on four hyperspectral image classification benchmark datasets.}
\label{tab:benchmarks}
\begin{tabular}{l|ccc|ccc|ccc|ccc} 
\toprule[1.2pt]
\multirow{2}{*}{Methods}                        & \multicolumn{3}{c|}{IP}                                 & \multicolumn{3}{c|}{KSC}                                & \multicolumn{3}{c|}{WHU-Hi-HanChuan}                    & \multicolumn{3}{c}{HyRANK-Dioni}                         \\ 
\cmidrule{2-13}
                                                & AA $\uparrow$   & OA $\uparrow$   & $\kappa$ $\uparrow$ & AA $\uparrow$   & OA $\uparrow$   & $\kappa$ $\uparrow$ & AA $\uparrow$   & OA $\uparrow$   & $\kappa$ $\uparrow$ & AA $\uparrow$   & OA $\uparrow$   & $\kappa$ $\uparrow$  \\ 
\midrule
SSTN~\cite{SSTN}               & 0.8982          & 0.9091          & 0.8961              & 0.8014          & 0.8770          & 0.8631              & 0.9281          & 0.9667          & 0.9610              & 0.9735          & 0.9687          & 0.9612               \\
SSSAN~\cite{SSSAN}             & 0.9218          & 0.9580          & 0.9521              & 0.8930          & 0.9313          & 0.9235              & 0.9664          & 0.9799          & 0.9764              & 0.9844          & 0.9854          & 0.9819               \\
CVSSN~\cite{CVSSN}             & 0.9731          & 0.9842          & 0.9820              & 0.9791          & 0.9852          & 0.9835              & 0.9820          & 0.9868          & 0.9846              & 0.9875          & \textbf{0.9883}          & \textbf{0.9855}      \\
MorphFormer~\cite{morphformer} & 0.8627          & 0.8266          & 0.8012              & 0.8644          & 0.9076          & 0.8972              & 0.9553          & 0.9769          & 0.9730              & 0.9510          & 0.9408          & 0.9262               \\
SSEFN~\cite{SSEFN}             & 0.9618          & 0.9821          & 0.9796              & 0.9834          & 0.9904          & 0.9893              & 0.9915          & 0.9937          & 0.9927              & 0.9871          & 0.9827          & 0.9785               \\
MambaHSI~\cite{MambaHSI}       & 0.9626          & 0.9737          & 0.9700              & 0.9854          & 0.9902          & 0.9891              & 0.9731          & 0.9822          & 0.9792              & 0.9876          & 0.9854          & 0.9819               \\
DSTC~\cite{DSTC_ECCV}         & 0.9875 & 0.9615          & 0.9857     & 0.9802          & 0.9860          & 0.9844              & 0.9885          & 0.9909          & 0.9894              & 0.9683          & 0.9633          & 0.9544               \\ 
S2Mamba~\cite{S2Mamba}         & 0.9791          & 0.9868 & 0.9850              & 0.9949          & 0.9976          & 0.9973              & 0.9946          & 0.9959          & 0.9953              & 0.9886          & 0.9859          & 0.9825               \\
\midrule
\rowcolor{Gray}
DSCC (Ours)   & \textbf{0.9877 }         & \textbf{0.9885}          & \textbf{0.9869}              & \textbf{0.9985} & \textbf{0.9983} & \textbf{0.9981}     & \textbf{0.9976} & \textbf{0.9975} & \textbf{0.9971}     & \textbf{0.9915} & 0.9878 & 0.9848               \\
\bottomrule[1.2pt]
\end{tabular}
\end{table*}

\begin{figure*}
    \centering
    \includegraphics[width=\textwidth]{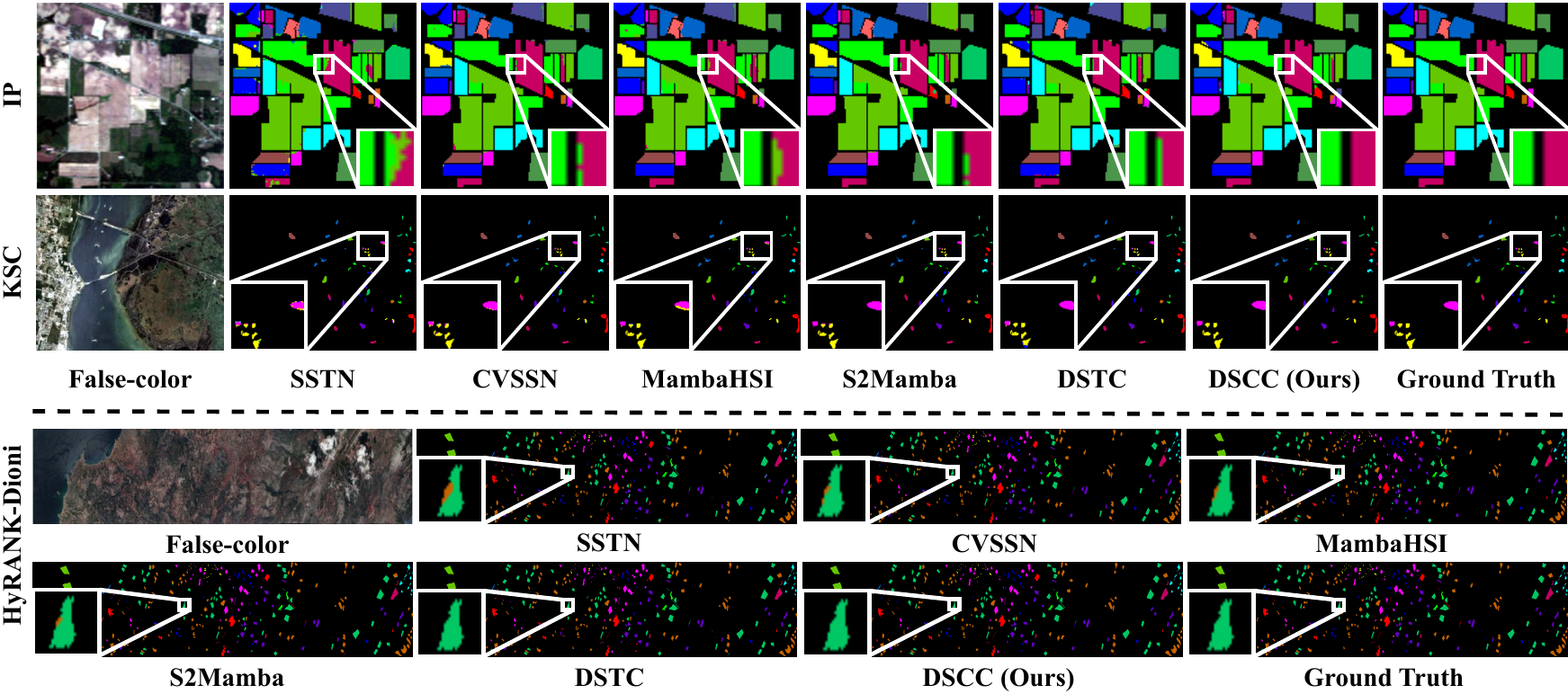}
    \caption{Qualitative results on IP, KSC and HyRANK-Dioni datasets. DSCC achieves the best classification performance with accurate boundary delineation.}
    \label{fig:three-datasets}
    \vspace{-4mm}
\end{figure*}

\begin{figure}
    \centering
    \includegraphics[width=\linewidth]{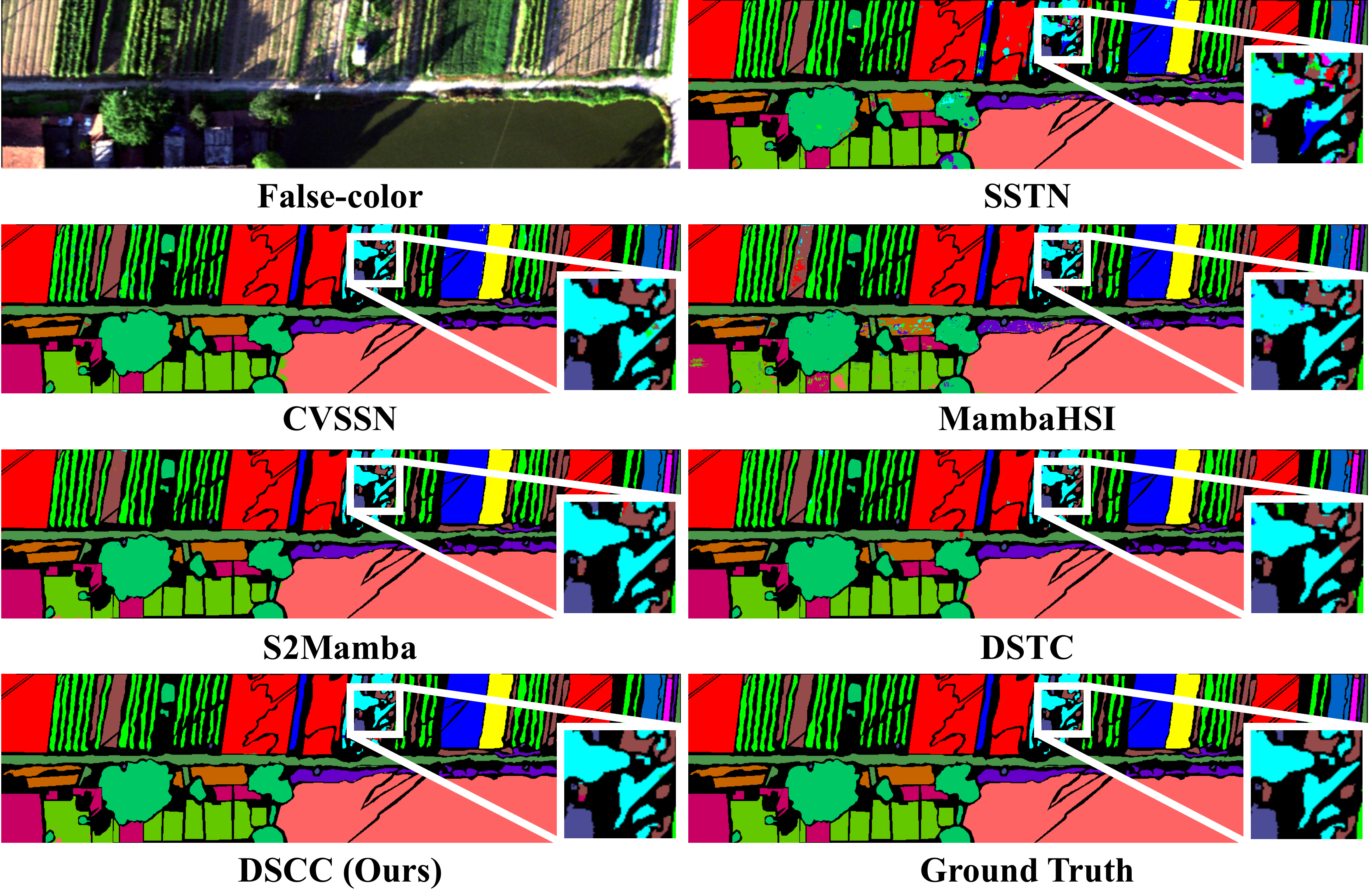}
    \caption{Qualitative results on the WHU-Hi-HanChuan dataset. Clustering similar pixels improves regional consistency and boundary accuracy in DSCC's output.}
    \label{fig:hanchuan}
\end{figure}

\smallsection{Data and Experimental Setups} 
The Indian Pines (IP) dataset comprises 145$\times$145 pixels with a spatial resolution of 20 m. It contains 200 spectral bands and represents 16 land cover classes with varying sample sizes.
The Kennedy Space Center (KSC) dataset covers an area of 512$\times$614 pixels and originally includes 176 spectral bands. After eliminating bands with low signal-to-noise ratios in the wavelength range of 400-2500 nm, the dataset is reduced to 5202 labeled samples across 13 classes.
The WHU-Hi-HanChuan dataset~\cite{WHU-Hi} consists of 1217$\times$303 pixels with a spatial resolution of approximately 0.109 m. It includes 274 spectral bands in the 400-1000 nm range, distributed across 16 land cover classes.
The HyRANK-Dioni dataset has a spatial dimension of 250$\times$1376 pixels, with 176 spectral bands and a spatial resolution of 30 m. It contains 14 land cover categories.

For experiments on these datasets, the hyperspectral images are cropped into non-overlapping 9$\times$9 patches. Unlike the WHU-OHS dataset, processing the entire image at once on these single-image benchmarks would lead to unfair comparisons with existing state-of-the-art methods that strictly adhere to the patch-based evaluation protocol. We adopt this standard patch-wise setup to ensure a rigorous and fair comparison.
Within each patch, Spectral-Consistent Pixel Aggregation modules group pixels into $\mathrm{M}_1 =$ 16 and $\mathrm{M}_2 =$ 8 spectral supertokens, respectively. A random split of 10\% and 90\% is used for training and testing. DSCC is trained with a batch size of 16, an initial learning rate of $1 \times 10^{-4}$, and 100 epochs, while DSTC uses a learning rate of $1 \times 10^{-3}$ with other settings kept the same. 
Final pixel-wise predictions are obtained by majority voting over all patches containing each pixel.

\smallsection{Quantitative Results}
The quantitative results on the benchmark datasets are summarized in~\cref{tab:benchmarks}. DSCC achieves strong and often leading performance across multiple evaluation metrics. It achieves the best results on all three metrics for the IP, KSC, and WHU-Hi-HanChuan datasets. For example, on the WHU-Hi-HanChuan dataset, DSCC achieves an OA of 0.9975, an AA of 0.9976, and a $\kappa$ of 0.9971, clearly outperforming S2Mamba, which attains 0.9959, 0.9946, and 0.9953, respectively. On the HyRANK-Dioni dataset, DSCC also performs exceptionally well, achieving the best AA and remaining highly competitive on OA and $\kappa$, falling short of the leading method by only 0.0005 and 0.0007.

DSCC consistently surpasses DSTC across all datasets, confirming the benefit of the proposed improvements. On the KSC dataset, for instance, DSCC achieves an AA of 0.9985, exceeding DSTC by 0.0183. A substantial gain is also observed on the IP dataset, where DSCC obtains an OA of 0.9885 compared to DSTC’s 0.9615, an improvement of 0.0270. These results comprehensively demonstrate the superiority of DSCC for hyperspectral image classification.

\smallsection{Qualitative Results}
Qualitative results are shown in~\cref{fig:three-datasets,fig:hanchuan}, with zoomed-in views for closer inspection. Both DSTC and DSCC leverage spectral similarity to aggregate pixels, yielding more accurate boundaries and more coherent spatial predictions compared with pixel-wise methods, as seen in the enlarged regions. By modeling global pixel-cluster correlations and refining center distributions, DSCC further enhances boundary preservation and classification accuracy over DSTC. Results on the IP dataset also confirm the effectiveness of these improvements. Compared with other open-source state-of-the-art methods, DSCC delivers clear advantages in both classification accuracy and boundary delineation, further validating its robustness and efficacy.

\begin{figure}
    \centering
    \includegraphics[width=1\linewidth]{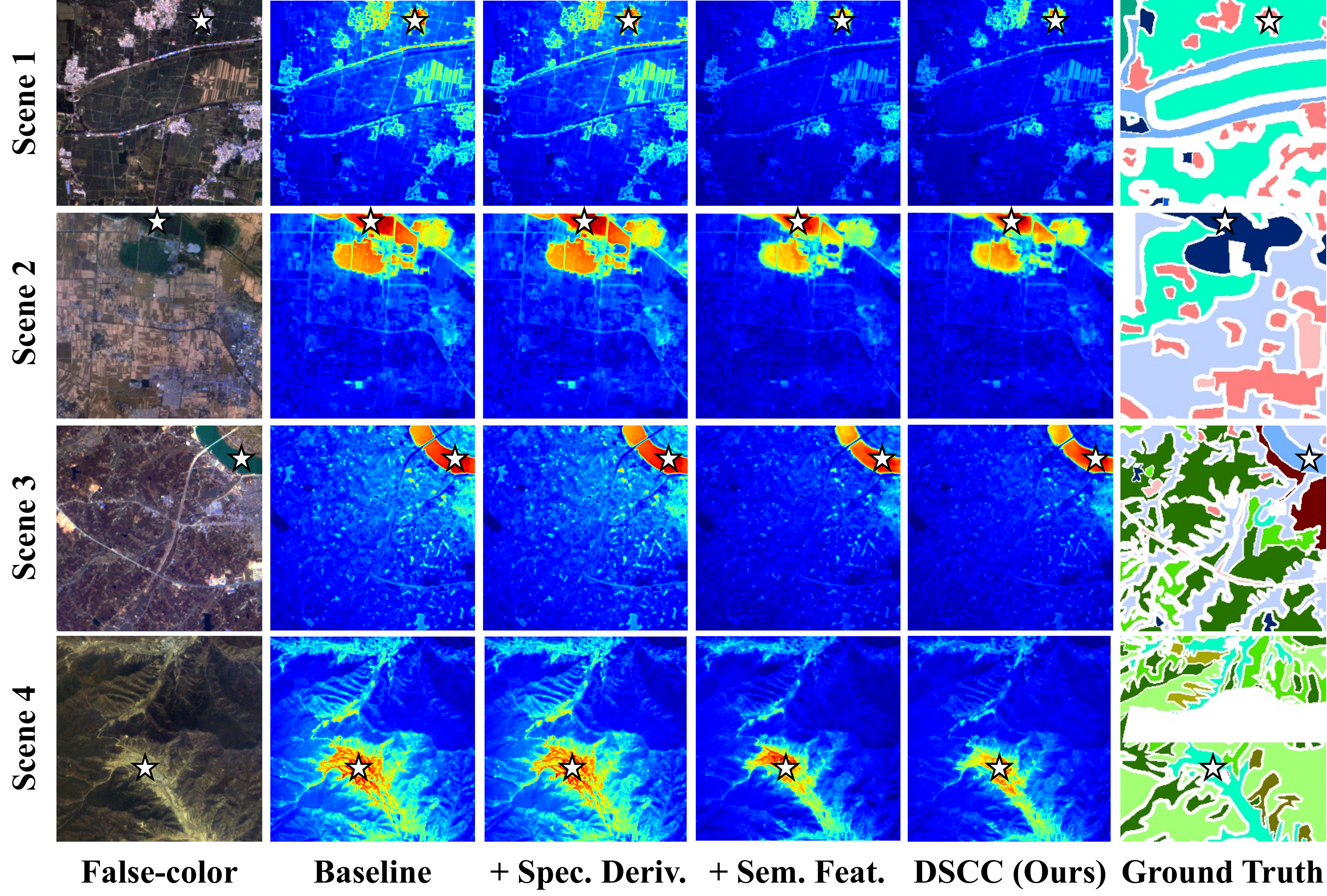}
    \caption{Ablation study of Spectral Derivative (Spec. Deriv.) and Semantic Feature (Sem. Feat.) on pixel-center similarity. White stars indicate selected centers. Red denotes higher similarity, and blue denotes lower similarity. }
    \label{fig:sim_ab}
\end{figure}

\subsection{Ablation Study}
We conduct ablation studies on the WHU-OHS dataset to evaluate the contribution of each component in DSCC.

\smallsection{Effect of Semantic Feature and Spectral Derivative}
As shown in~\cref{tab:ablation_SF_SD}, we examine the impact of semantic features and spectral derivative on classification performance. Compared with the baseline, adding semantic features alone improves OA to 0.801 and CF1 to 0.725, while adding only the spectral derivative increases OA to 0.794 and CF1 to 0.720. The larger gain from semantic features indicates their stronger role in pixel clustering and center filtering. When both components are combined, all metrics reach their highest values, showing that semantic features and spectral derivative complement each other.

\begin{table}[tp]
    \centering
    \caption{Effect of Semantic Feature (SF) and Spectral Derivative (SD).}
    \label{tab:ablation_SF_SD}
    \setlength{\tabcolsep}{10pt}
    \begin{tabular}{cc|cccc}
    \toprule[1.2pt]
        SF & SD & CF1 $\uparrow$ & OA $\uparrow$ & $\kappa$ $\uparrow$ & mIoU $\uparrow$ \\
    \midrule
        \nohave & \nohave & 0.713 & 0.791 & 0.765 & 0.582 \\ 
        \nohave & \have & 0.720 & 0.794 & 0.768 & 0.590 \\ 
        \have & \nohave & 0.725 & 0.801 & 0.775 & 0.597 \\ 
        \have & \have & \textbf{0.728} & \textbf{0.802} & \textbf{0.776}  & \textbf{0.602} \\
    \bottomrule[1.2pt]
    \end{tabular}
\end{table}

\begin{table}[tp]
    \centering
    \caption{Comparison of Supervision Strategies.}
    \label{tab:CPSL}
    \setlength{\tabcolsep}{10pt}
    \begin{tabular}{l|cccc}
    \toprule[1.2pt]
        Supervision & CF1 $\uparrow$ & OA $\uparrow$ & $\kappa$ $\uparrow$ & mIoU $\uparrow$ \\
    \midrule
        Hard label & 0.594  & 0.693 & 0.651 & 0.450 \\ 
        Dense-CE &  0.716 & 0.795 & 0.768 & 0.589 \\ 
        \rowcolor{Gray}
        Soft label & \textbf{0.728} & \textbf{0.802} & \textbf{0.776}  & \textbf{0.602} \\
    \bottomrule[1.2pt]
    \end{tabular}
\end{table}

\begin{figure}
    \centering
    \includegraphics[width=1\linewidth]{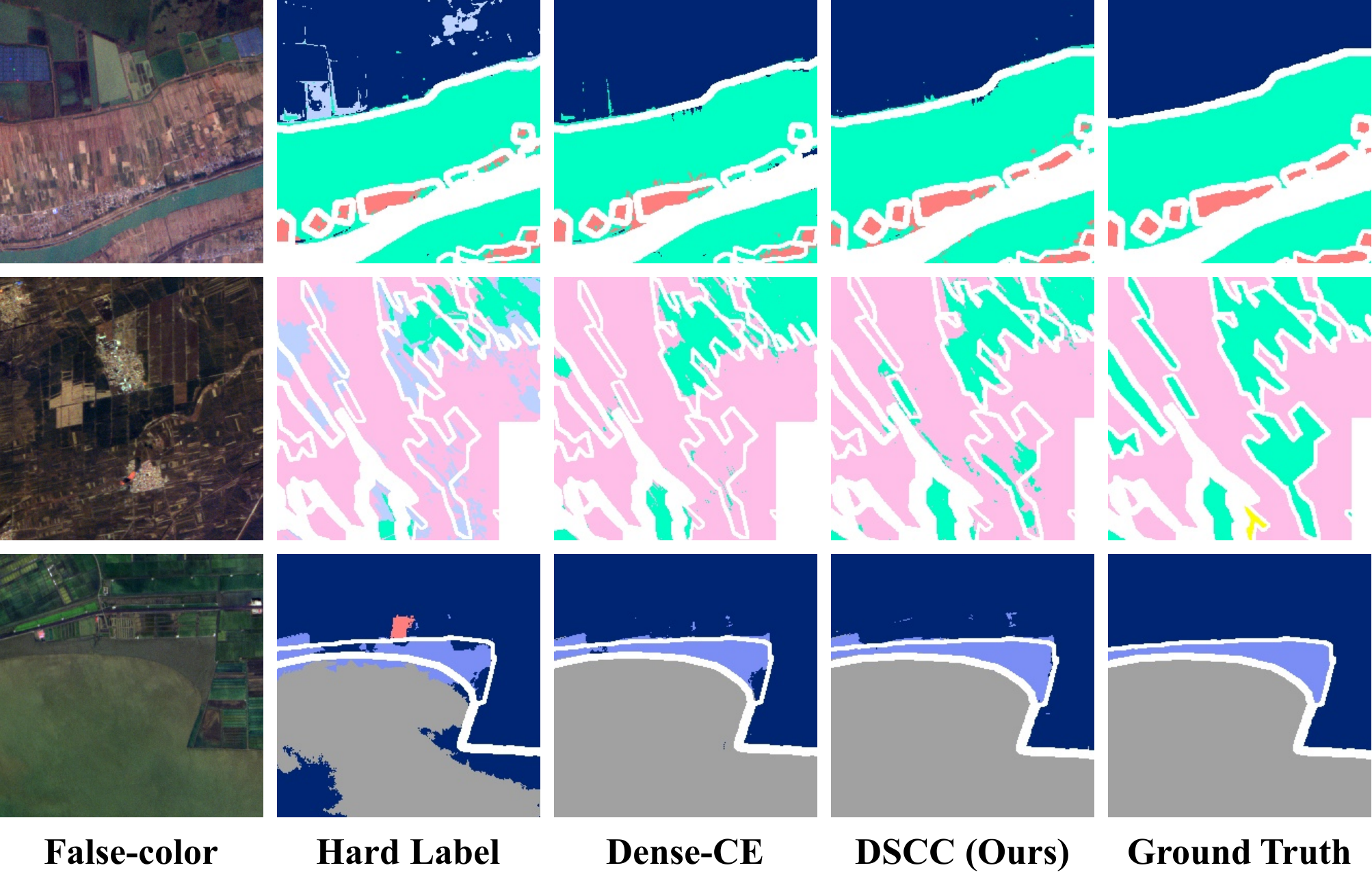}
    \caption{Comparison of classification results under different supervision strategies. Soft label used in our DSCC achieves the best performance, underscoring its effectiveness.}
    \label{fig:ab_supervis}
\end{figure}

\cref{fig:sim_ab} visualizes similarity heatmaps between selected centers and their neighboring pixels under different configurations. The baseline groups spectrally similar pixels reasonably well but still shows moderate responses to dissimilar ones, revealing limited discrimination. Adding only the spectral derivative yields little improvement. In contrast, semantic features substantially suppress responses to dissimilar pixels and sharpen focus on similar ones. Combining semantic features and spectral derivative further enhances discrimination, consistent with the quantitative results.

\smallsection{Comparison of Supervision Strategies}
To assess the effectiveness of the Category Proportion-aware Soft Label, we compare it with two alternatives: a hard label assigning each supertoken to a single class and dense cross-entropy (dense-CE), which projects supertokens back to the pixel space for pixel-level supervision. As shown in~\cref{tab:CPSL}, hard labels produce the weakest results, likely due to amplified label noise from one-hot assignments. Dense-CE performs better, but our soft-label supervision achieves the best performance, reaching a CF1 of 0.728, OA of 0.802, $\kappa$ of 0.776, and mIoU of 0.602. The visual comparisons in~\cref{fig:ab_supervis} show that hard labels degrade segmentation quality, while the proposed CPSL yields the most consistent predictions, confirming its robustness to mixed-class supertokens and its stabilizing effect during training.

\begin{table}[tp]
    \centering
    \caption{Ablation Study on Number of Layers in UNet.}
    \label{tab:unet_ablation}
    \setlength{\tabcolsep}{6pt}
    \begin{tabular}{c|cccc}
        \toprule[1.2pt]
        Layers & CF1 $\uparrow$ & FLOPs (G) & \#Param (M) & Speed (FPS) \\
        \midrule
        1 & 0.719 & 23.92 & 7.28 & 312.62 \\
        \rowcolor{Gray}
        2 & 0.728 & 34.33 & 8.57 & 197.75 \\
        3 & 0.721 & 44.70 & 13.75 & 137.01 \\
        4 & 0.721 & 55.05 & 34.47 & 110.83 \\
        \bottomrule[1.2pt]
    \end{tabular}
\end{table}

\begin{table}[tp]
\centering
\caption{Effect of spatial mask size in Spectral-Consistent Pixel Aggregation.}
\label{tab:ablation_mask_size}
\setlength{\tabcolsep}{10pt}
    \begin{tabular}{c|cccc}
    \toprule[1.2pt]
    \textbf{Mask Size} & CF1 $\uparrow$ & OA $\uparrow$ & $\kappa$ $\uparrow$ & mIoU $\uparrow$ \\
    \midrule
    5 & 0.725 & 0.797 & 0.771 & 0.597 \\
    7 & 0.725 & 0.797 & 0.772 & 0.596 \\
    \rowcolor{Gray}
    9 & \textbf{0.728} & \textbf{0.802} & \textbf{0.776} & \textbf{0.602} \\
    11 & 0.724 & 0.797 & 0.771 & 0.594 \\
    13 & 0.715 & 0.796 & 0.770 & 0.586 \\
    \bottomrule[1.2pt]
    \end{tabular}
\end{table}

\begin{table}[h]
\centering
\caption{Effect of $\mathrm{K}$ value in Density-Isolation Center Filtering.}
\label{tab:ablation_k_value}
\setlength{\tabcolsep}{13pt}
    \begin{tabular}{c|cccc}
    \toprule[1.2pt]
    $\mathrm{K}$ & CF1 $\uparrow$ & OA $\uparrow$ & $\kappa$ $\uparrow$ & mIoU $\uparrow$ \\
    \midrule
    5 & 0.725 & 0.797 & 0.771 & 0.597 \\
    7 & 0.725 & 0.797 & 0.772 & 0.596 \\
    \rowcolor{Gray}
    9 & \textbf{0.728} & \textbf{0.801} & \textbf{0.776} & \textbf{0.602} \\
    11 & 0.724 & 0.797 & 0.771 & 0.594 \\
    13 & 0.715 & 0.796 & 0.770 & 0.586 \\
    \bottomrule[1.2pt]
    \end{tabular}
\end{table}

\smallsection{Effect of UNet Layers}
\cref{tab:unet_ablation} analyzes the impact of the number of downsampling and upsampling layers in the UNet. Network depth significantly affects performance and efficiency. A one-layer UNet is the most efficient (7.28~M parameters, 312.62~FPS) but achieves a relatively low CF1 of 0.719. The two-layer UNet delivers the best classification performance (CF1: 0.728) with acceptable computational cost. Increasing depth further leads to substantial overhead with no accuracy gains. This suggests that shallow semantic features are more effective for pixel aggregation and center filtering. Thus, we adopt the two-layer UNet as the default architecture.

\smallsection{Effect of Spatial Mask Size}
To study the influence of spatial locality in the Spectral-Consistent Pixel Aggregation module, we evaluate different mask sizes, as summarized in~\cref{tab:ablation_mask_size}. The mask controls which centers each pixel can associate with. A mask size of 9 yields the highest accuracy. Smaller masks restrict the local neighborhood too strongly, reducing expressive capacity, while larger masks introduce spatially irrelevant centers that weaken local consistency. We therefore set the mask size to 9 in our implementation.

\smallsection{Effect of K in Center Filtering}
We also investigate the effect of the hyperparameter $\mathrm{K}$ in the Density-Isolation Center Filtering module, which controls the locality of density estimation. As shown in~\cref{tab:ablation_k_value}, using 9 neighbors strikes the best balance between local sensitivity and contextual awareness. In practice, $\mathrm{K}$ does not need to be tuned over a wide range for each dataset. Extremely small $\mathrm{K}$ leads to unstable density estimation, while overly large $\mathrm{K}$ over-smooths local structure. Therefore, $\mathrm{K}$ is naturally upper-bounded by the number of available centers and is preferably chosen as a moderate neighborhood size.

\begin{table}[tp]
    \centering
    \caption{Ablation Study on Transformer and Mamba Integration Strategies.}
    \setlength{\tabcolsep}{6pt}
    \label{tab:transformer_mamba}
    \begin{tabular}{l|cccc}
        \toprule[1.2pt]
        Strategy & CF1 $\uparrow$ & OA $\uparrow$ & $\kappa$ $\uparrow$ & mIoU $\uparrow$ \\
        \midrule
        Transformer Only & 0.724 & 0.793 & 0.767 & 0.595 \\
        \rowcolor{Gray}
        Transformer + Mamba & \textbf{0.728} & \textbf{0.802} & \textbf{0.776}  & \textbf{0.602} \\
        Mamba + Transformer & 0.719 & 0.794 & 0.768 & 0.590 \\
        \bottomrule[1.2pt]
    \end{tabular}
\end{table}

\begin{table}[tp]
    \centering
    \caption{Ablation Study on Clustering Center Filtering Patterns.}
    \setlength{\tabcolsep}{10pt}
    \label{tab:Filtering}
    \begin{tabular}{cc|cccc}
    \toprule[1.2pt]
        $\mathrm{M}_1$  & $\mathrm{M}_2$ & CF1 $\uparrow$ & OA $\uparrow$  & $\kappa$ $\uparrow$ & mIoU $\uparrow$ \\
    \midrule
        144 & 144 & 0.718 & 0.793 & 0.767 & 0.588  \\
        144 & 72  & 0.709 & 0.787 & 0.760 & 0.578  \\
        144 & 36  & 0.704 & 0.780 & 0.752 & 0.573  \\
        256 & 256 & 0.724 & 0.801 & 0.776 & 0.595  \\
        \rowcolor{Gray}
        256 & 128 & \textbf{0.728} & \textbf{0.802} & \textbf{0.776} & \textbf{0.602}  \\
        256 & 64  & 0.714 & 0.789 & 0.762 & 0.583  \\
        400 & 400 & 0.719 & 0.800 & 0.775 & 0.591  \\
        400 & 200 & 0.713 & 0.796 & 0.770 & 0.582  \\
        400 & 100 & 0.716 & 0.795 & 0.768 & 0.586 \\
    \bottomrule[1.2pt]
    \end{tabular}
\end{table}

\smallsection{Effect of Transformer-Mamba Integration Strategies}
We evaluate different integration strategies for the Transformer and Mamba modules, as reported in~\cref{tab:transformer_mamba}. “Transformer Only’’ uses four stacked Transformer blocks. “Transformer + Mamba’’ alternates the two modules starting with a Transformer block, while “Mamba + Transformer’’ starts with a Mamba block. The “Transformer + Mamba’’ configuration achieves the best results, demonstrating the effectiveness of the hybrid design. A pure Mamba architecture fails to converge, likely due to its sensitivity to initialization and implementation details~\cite{pmlr-v202-fu23a}. Considering both stability and accuracy, we adopt the “Transformer + Mamba’’ structure.

\smallsection{Effect of Center Filtering Patterns}
As shown in~\cref{tab:Filtering}, the center filtering scheme substantially affects classification performance. $\mathrm{M}_2$ controls the compactness of the filtered center set: using too many centers (e.g., 256→256 or 400→400) introduces redundancy and weakens feature discrimination, whereas using too few (e.g., 144→72 or 256→64) leads to sparse spatial distribution and loss of small-object details. The 256→128 configuration achieves the best balance between representational compactness and spatial coherence, yielding the highest CF1, OA, and mIoU scores. Therefore, in practice, $\mathrm{M}_2$ can be selected as a moderate fraction of $\mathrm{M}_1$ rather than re-designed from scratch for each scenario, with the natural constraint $\mathrm{M}_2 \leq \mathrm{M}_1$.

\section{Conclusion}
In this work, we propose a novel dual-stage clustering-based classifier, DSCC, for hyperspectral image classification. The method uses a multi-layer architecture that first aggregates spectrally similar pixels into supertokens via Spectral-Consistent Pixel Aggregation, and then applies Density-Isolation Center Filtering to refine cluster centers and improve robustness to scale variation. We further introduce a Category Proportion-aware Soft Label mechanism to incorporate category distribution information into learning. Extensive experiments show that DSCC achieves state-of-the-art accuracy and a superior trade-off in computational efficiency, indicating that DSCC is particularly promising for high-spatial-resolution hyperspectral scenarios, where preserving fine boundaries and maintaining regional consistency are both critical. Ablation studies validate the effectiveness of each component.

Future research may focus on adaptive determination of the neighborhood size for density estimation and the numbers of clustering centers, in order to improve flexibility across scenes with different spatial structures. Another promising direction is to develop more efficient global association schemes for ultra-large hyperspectral images, further improving scalability while preserving boundary accuracy and regional consistency.

\section*{Acknowledgments}
This work was financially supported by the National Natural Science Foundation of China (No. 62101032), the Chongqing Excellent Young Scientists Fund (No. CSTB2025NSCQ-JQX0017), and the High-Quality Development Special Project of the Ministry of Industry and Information Technology (TC240HAJ9-35).



\bibliographystyle{cas-model2-names}

\bibliography{cas-refs}

\end{document}